\documentclass[runningheads]{llncs}

\usepackage{eccv}

\usepackage{eccvabbrv}

\usepackage{multirow}
\usepackage{graphicx}
\usepackage{booktabs}

\usepackage[accsupp]{axessibility}  

\usepackage{hyperref}

\usepackage{orcidlink}

\begin{document}

\title{Fast Point Cloud Geometry Compression \\
with Context-based Residual Coding \\
and INR-based Refinement} 

\titlerunning{Point cloud geometry compression with CRCIR}

\author{Hao Xu\inst{1}\orcidlink{0000-0001-5685-5225} \and
Xi Zhang\inst{2}\orcidlink{0000-0002-1993-6031} \and
Xiaolin Wu\inst{1}\thanks{Corresponding author.}\orcidlink{0000-0002-0103-5374}}

\authorrunning{H. Xu et al.}

\institute{McMaster University \and
Shanghai Jiao Tong University\\
\email{xu338@mcmaster.ca,~xzhang9308@gmail.com,~xwu@ece.mcmaster.ca}}
\maketitle
\begin{abstract}
  Compressing a set of unordered points is far more challenging than compressing images/videos of regular sample grids, because of the difficulties in characterizing neighboring relations in an irregular layout of points. 
  Many researchers resort to voxelization to introduce regularity, but this approach suffers from quantization loss.
  In this research, we use the KNN method to determine the neighborhoods of raw surface points.
  This gives us a means to determine the spatial context in which the latent features of 3D points are compressed by arithmetic coding.  
  As such, the conditional probability model is adaptive to local geometry, leading to significant rate reduction.  
  Additionally, we propose a dual-layer architecture where a non-learning base layer reconstructs the main structures of the point cloud at low complexity, while a learned refinement layer focuses on preserving fine details. This design leads to reductions in model complexity and coding latency by two orders of magnitude compared to SOTA methods.
  Moreover, we incorporate an implicit neural representation (INR) into the refinement layer, allowing the decoder to sample points on the underlying surface at arbitrary densities. 
  This work is the first to effectively exploit content-aware local contexts for compressing irregular raw point clouds, achieving high rate-distortion performance, low complexity, and the ability to function as an arbitrary-scale upsampling network simultaneously. Our code is available \href{https://github.com/hxu160/CRCIR_for_PCGC}{here}.
  
\end{abstract}

\section{Introduction}
\label{sec:intro}
Point cloud is a common 3D representation in many applications, such as virtual reality, robotics and autonomous driving. Compared with 2D images, 3D point clouds generate much greater volume of data, and thus need to be
compressed for efficient storage and transmission. Point cloud compression poses a technical challenge in coding 3D points in unordered and irregular configurations. Although methods employing regular structures like octrees~\cite{huang2020octsqueeze, que2021voxelcontext, fu2022octattention, song2023efficient} or voxel grids~\cite{wang2021lossy, wang2021multiscale, wang2022sparse, pang2022grasp,zhang2023yoga} are available, they have relatively low coding efficiency in compressing high-precision coordinates.  
Other methods directly compress raw point clouds without the voxelization. They typically adopt an autoencoder architecture and entropy code point-wise latent features~\cite{yan2019deep,huang20193d,you2021patch,wiesmann2021deep,you2022ipdae,he2022density,huang20223qnet}. 
Unlike in the entropy coding of image/video, it is difficult to characterize neighboring relations in context-based arithmetic coding. 
Two straightforward methods are proposed to circumvent the issue: global maximum pooling~\cite{yan2019deep,huang20193d} and downsampling~\cite{you2021patch,wiesmann2021deep,you2022ipdae,he2022density,huang20223qnet}. 
These methods suffer from three common drawbacks: 1. inability to form local coding contexts due to irregularity of point clouds, reducing coding efficiency; 2. non-scalability with a fixed cardinality of the reconstructed point set, reducing flexibility for downstream tasks; 3. inability to balance the trade-off between rate-distortion performance and computational complexities, reducing practicability in real applications.

To overcome the limitations of the existing methods, we propose a novel method of 
Context-based Residual Coding and Implicit neural representation (INR) based Refinement,
denoted by CRCIR.  It consists of two compression layers: a non-learning base layer for main structure reconstruction and a learned refinement layer for recovering finer details.  
First, the base layer builds a coarse geometry representation with very low complexity. Then in the refinement layer,
we use the K-Nearest Neighbors (KNN) to characterize neighboring relations for exploiting correlations between latent features in a locality.  The CRCIR method allows us to
construct a content-adaptive conditional entropy model that removes statistical redundancies among neighboring features, leading to superior rate-distortion performance.  It is worth noting that our method achieves both lower complexity and better rate-distortion performance than the existing methods.

Furthermore, unlike in existing methods the decoder directly transforms latent features into a 3D point set of a fixed cardinality, our decoder allows flexible point sampling on a learned implicit 3D surface at arbitrary densities. This is achieved by integrating INR into the decoder of the refinement layer. 
In addition, the CRCIR decoder is conditioned on latent features extracted by the CRCIR encoder and maps the coordinates of any 3D point to the offset between it and the underlying surface. Additionally, incorporating INR into the refinement layer eliminates the need to retrain INR on a per-object basis~\cite{dupont2021coin,dupont2022coin++,strumpler2022implicit,postels20233d}, streamlining the compression process. To our knowledge, this paper is the first to combine the advantages of both INR and non-linear transform coding (NTC) in point cloud geometry compression.  The above design strategy may be generalized to utilize INR in data compression of other modalities. 

The contributions of our paper are summarized as follows: 
\begin{itemize}
    \item We design a novel CRCIR method for point cloud geometry compression. It achieves not only superior rate-distortion performance but also the reductions of model complexity and coding latency. Both reductions are by two orders of magnitude compared to the SOTA method 3QNet~\cite{huang20223qnet}.
    \item We propose a novel conditional entropy model for the compression of latent features, being the first to effectively form local contexts of adaptive arithmetic coding, against the irregularity of raw 3D point clouds.
    \item We incorporate INR into the refinement layer of the CRCIR compression system, allowing the decoder to sample points on the learned implicit surface at arbitray densities.
\end{itemize}
\section{Related Work}
\subsection{Deep compression of point cloud geometry}
Recently, deep learning has significantly influenced the research on point cloud compression. One facet of this research endeavor involves utilizing deep conditional entropy models to losslessly encode the octree-structured point cloud geometry into a compact bitstream~\cite{huang2020octsqueeze, que2021voxelcontext, fu2022octattention, song2023efficient}. Moreover, parallel to the advancements in deep lossy image compression~\cite{balle2016end, balle2018variational, minnen2018joint, cheng2020learned, guo2020deep, zhang2021attention, he2022elic, kim2022joint, zhang2023lvqac, liu2023learned}, some researchers have shifted to using a combination of autoencoder architecture and deep entropy models to optimize the rate-distortion trade-off in an end-to-end manner. The techniques can be further categorized into voxel-based~\cite{wang2021lossy, wang2021multiscale, wang2022sparse, pang2022grasp,zhang2023yoga} and point-based methods~\cite{yan2019deep, huang20193d, wiesmann2021deep, he2022density, you2021patch,you2022ipdae,huang20223qnet}. The former needs to voxelize the input before feeding it into stacked 3D convolutional layers, and the voxelization process results in a critical artifact of missing points in the reconstructed point clouds. In contrast, the latter preserves the original density and details through direct processing of raw point clouds. 

Point-based methods struggle to efficiently reduce redundancy due to the irregular and unordered nature of raw point clouds. To reduce the bitrate, two straightforward methods are proposed: global maximum pooling~\cite{yan2019deep,huang20193d} and downsampling~\cite{you2021patch,wiesmann2021deep,you2022ipdae,he2022density,huang20223qnet}. Global maximum pooling, although simple to implement, is only suitable for sparse point clouds, as it faces challenges in reconstructing a dense point cloud from a single feature vector. On the other hand, the second strategy is more effective for dense point clouds because a feature vector is only responsible for reconstructing a specific patch of points. Among them, 3QNet~\cite{huang20223qnet} utilizes a learned codebook for vector quantization of latent features, while others~\cite{you2021patch,you2022ipdae,he2022density} use scalar quantization. 
\subsection{Point cloud upsampling}
Point cloud upsampling aims to transform a sparse point cloud into a dense and uniformly distributed set of points, similar to how image super-resolution increases image resolution~\cite{dong2015image,shi2016real,hu2019meta,saharia2022image,guo2022data,luo2021functional,luo2024and}.
Previous methods typically expand the number of points by converting expanded features to coordinates or offsets~\cite{yu2018pu, yifan2019patch, li2019pu, qian2021pu, li2021point}, necessitating a model trained for a specific upsampling rate.
Recently, magnification-flexible point cloud upsampling methods have become more popular due to the increased flexibility  ~\cite{ye2021meta,qian2021deep,feng2022neural,zhao2022self,he2023grad}. 
Among them, INR has been leveraged for its ability to naturally support arbitrary-scale upsampling~\cite{feng2022neural,zhao2022self,he2023grad}. However, such methods suffer from computational inefficiency. Specifically, Grad-PU~\cite{he2023grad} must iteratively refine initialized points, requiring both forward and compute-intensive backward propagation at each iteration to determine the refinement step and direction respectively.
\begin{figure}
    \centering
    \includegraphics[width=\linewidth]{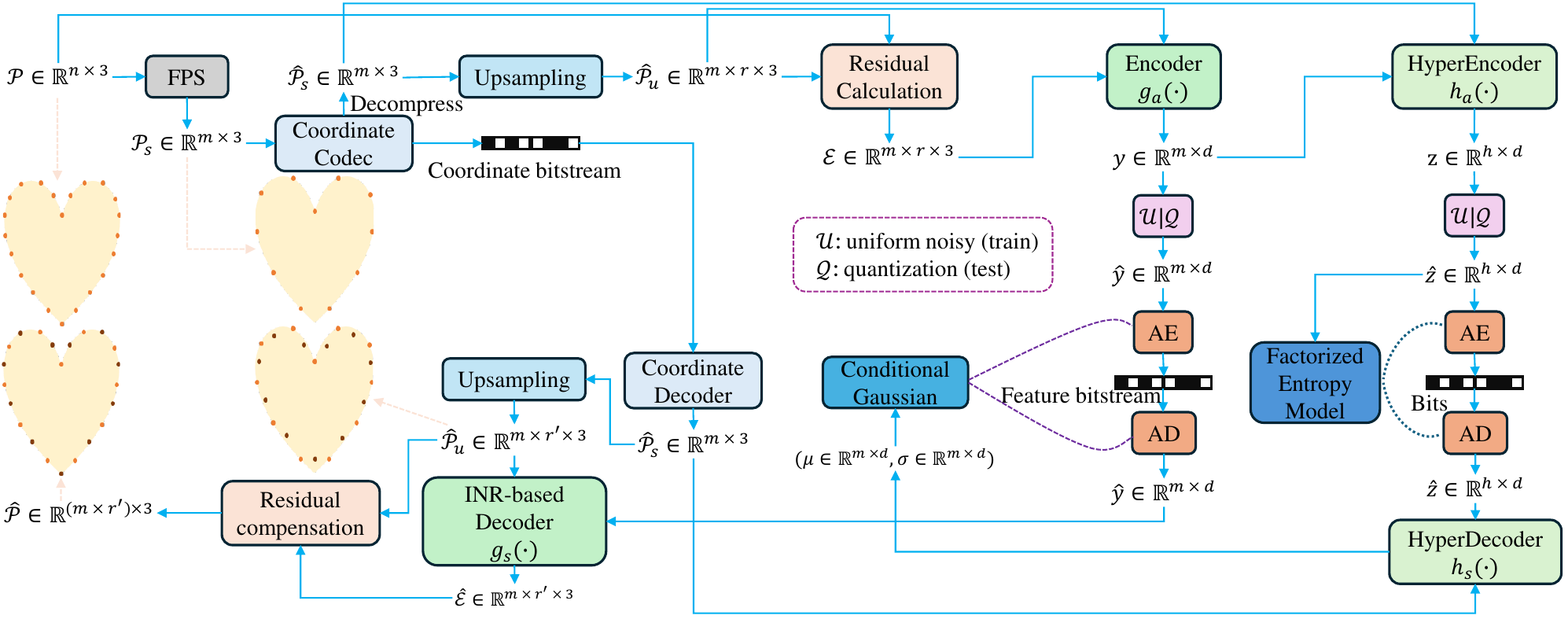}
    \caption{The overall architecture of the proposed CRCIR point cloud geometry compression system. 
    The base layer contains a pair of non-learning downsampling and upsampling modules and a traditional point cloud codec Google Draco~\cite{galligan2018google}. It produces the base layer point cloud $\hat{\mathcal{P}}_u$ in a "downsampling$\to$compressing$\to$upsampling" manner. 
    The refinement layer compress the residuals by learning a graph-based content-aware entropy model conditioned on the local context and base layer. The encoder $g_a(\cdot)$, decoder $g_s(\cdot)$, and hyperencoder $h_a(\cdot)$, hyperdecoder $h_s(\cdot)$ are jointly trained with the entropy model in terms of rate-distortion optimization.}
    \label{fig:system}
\end{figure}
\section{Method}
In this section, we will introduce the proposed CRCIR point cloud geometry compression system and its two key technical contributions: (1) an INR-based refinement layer for flexible point sampling on a learned implicit 3D surface at arbitrary densities; (2) a novel graph-based, content-adaptive conditional entropy model for the compression of unordered and irregular point clouds; 
\subsection{Overview}
The overall architecture of the proposed CRCIR compression system is presented in \cref{fig:system}. It consists of a low complexity base layer for reconstructing main structures and an INR-based refinement layer for further recovering finer details. Given a dense 3D point cloud $\mathcal{P}\in\mathbb{R}^{n\times3}$ to be compressed, the compression workflow involves the following steps:
\begin{enumerate}
    \item $\mathcal{P}\in\mathbb{R}^{n\times3}$ is first downsampled to $\mathcal{P}_s\in\mathbb{R}^{m\times3}$ using the farthest point sampling (FPS) method~\cite{qi2017pointnet++} and then compressed by an efficient traditional coordinate codec Google Draco~\cite{galligan2018google}. 
    The decompressed point cloud is denoted as $\hat{\mathcal{P}}_s$ and then fed into a non-learning upsampling module to obtain a dense yet coarse point cloud $\hat{\mathcal{P}}_u$ as the base layer point cloud.

    \item The residuals $\mathcal{E}$ between $\mathcal{P}$ and $\hat{\mathcal{P}}_u$ are transformed and quantized to latent features $\hat{y}$ and then compressed using a learned NTC method as the refinement layer. 
    Specifically, to improve the coding efficiency, we learn a conditional entropy model based on the downsampled point cloud $\hat{\mathcal{P}}_s$ for compressing the residuals $\mathcal{E}$.

    \item In the decoder, the decompressed residuals $\hat{\mathcal{E}}$ are predicted based on the compressed features $\hat{y}$ and the base layer point cloud $\hat{\mathcal{P}}_u$. After that, the base layer point cloud $\hat{\mathcal{P}}_u$ is refined by adding the decompressed residuals $\hat{\mathcal{E}}$, resulting in a finer reconstruction $\hat{\mathcal{P}}$ with higher fidelity.
\end{enumerate}
It's noteworthy that an INR is incorporated into the decoder of the refinement layer. This INR-based refinement layer can take as input the coordinates of arbitrary density and refine the input points by predicting the residuals between the input and the underlying surface represented by a dense point set. 
That is to say, by adjusting the decoder-end upsampling rate $r^{\prime}$ to an arbitrary value (same with or different from the encoder-end upsampling rate $r$), we can control the density of the reconstructed point cloud to any desired level, not necessarily the same as the original point cloud. This design enables our compression system to function as an arbitrary-scale point cloud reconstruction network.
\subsection{Low complexity base layer}
In point clouds, most points belong to the smooth regions, also known as low-frequency components. These areas make up the main structures of point cloud and generally are easier to reconstruct. On the other hand, sharp regions, characterized by a minority of points and known as high-frequency components, present challenges in achieving accurate reconstruction. 
Existing methods for compressing point clouds often rely on complex neural networks to handle all points, including those in both smooth and sharp regions. However, this approach leads to inefficient use of computational resources, especially when reconstructing smooth regions. To address this issue, we design a base layer dedicated to reconstructing main structures with very low complexity. 
Specifically, the downsampling and compression module in the base layer both adopts the efficient off-the-shelf algorithms FPS~\cite{qi2017pointnet++} and Google Draco~\cite{galligan2018google}, respectively. To obtain the base layer point cloud $\hat{\mathcal{P}}_u$, we design a simple but effective upsampling method to predict a dense point cloud $\hat{\mathcal{P}}_u$ from the downsampled and decompressed point cloud $\hat{\mathcal{P}}_s$, 
which is defined as: 
\begin{equation}
    \hat{p}_u^{(ij)}=\hat{p}_s^{(i)}+u_{ij}(\hat{p}_s^{(ij)}-\hat{p}_s^{(i)}),~1\leq i\leq m,~1\leq j\leq r.
    \label{eq:interp}
\end{equation}
Here $\hat{p}_s^{(i)}$ represents the $i$-th point in $\hat{\mathcal{P}}_s$, the set $\{\hat{p}_s^{(ij)}|1\leq j\leq r\}$ comprises the nearest $R$ neighbors of $\hat{p}_s^{(i)}$ in $\hat{\mathcal{P}}_s$, and $u_{ij}$ denotes the associated interpolation weights. The selection of the interpolation weights are provided in the supplementary material. 
\begin{figure}
	\centering
	\includegraphics[width=\linewidth]{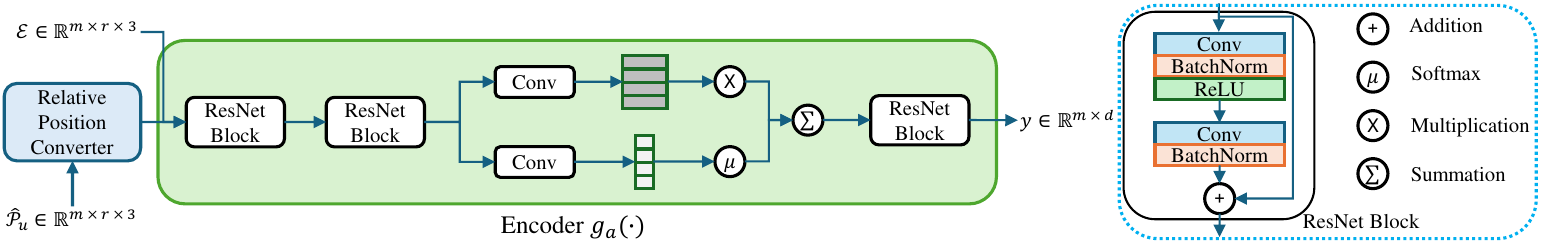}
	\caption{The architecture of the encoder $g_a(\cdot)$. It extracts a $d$-dimensional feature from all residuals in a cluster by an attention-based weighting block~\cite{Boulch_2022_CVPR} for further encoding. In the weighting block, one branch re-encodes features while the other branch learns content-adaptive weights followed by softmax normalization. 
    }
	\label{fig:encoder}
\end{figure}

\subsection{INR-based refinement layer} 
Point cloud reconstruction can be regarded as sampling points on the underlying surface. However, existing point-based compression methods~\cite{yan2019deep, huang20193d, wiesmann2021deep, he2022density, you2021patch,huang20223qnet} often neglect the modeling of the underlying surface and simply learn a mapping from latent features to the coordinates at a predefined density. The absence of surface modeling prevents these methods from flexibly controlling the density of the reconstructed point cloud. 

To address this issue, we propose an INR-based decoder for the refinement layer. 
This INR-based decoder take the coordinates of any points as input and predicts the residuals between these points and the underlying surface. In this context, the set of points whose residuals are $\mathop{0}\limits ^{\rightarrow}$ implicitly represent the underlying surface. Sampling points on this learned implicit surface involves three steps: 
1. sampling initialized points within the bounding box; 2. predicting residuals for the initialized points; 3. adding the predicted residuals to the initialized points. 
Specifically, the step 1 is achieved by the base layer and the sampling density is determined by the decoder-end base layer upsampling rate $r^{\prime}$. By setting $r^{\prime}$ to a desired value, it becomes possible to sample a set of points on the underlying surface with the intended target cardinality. Besides, $r^{\prime}$ can exceed the encoder-end base layer upsampling rate $r$. In this case, a denser point cloud can be reconstructed compared to the original one. 

Furthermore, to eliminate the need for retraining INR on a per-object basis in decoder-only architecture, the proposed method pairs the INR-based decoder $g_s(\cdot)$ with an encoder $g_a(\cdot)$. The decoder $g_s(\cdot)$ employs a similar architecture to the decoder in~\cite{peng2020convolutional} and the architecture of the encoder is shown in \cref{fig:encoder}. Instance-specific information is captured by the encoder $g_a(\cdot)$ and encoded into learned latent features, which then condition the shared decoder $g_s(\cdot)$. This design allows the proposed CRCIR method to streamline the compression process, significantly reducing coding latency.

The overall refinement layer operates by fusing the residuals $\mathcal{E}$ of the base layer point cloud $\hat{\mathcal{P}}_u$ into point-wise latent features with the encoder $g_a(\cdot)$ and predicting residuals with the decoder $g_s(\cdot)$ based on these features. To permit parallel computation, we partition $\hat{\mathcal{P}}_u$ into several clusters and let the encoder $g_a(\cdot)$ and the decoder $g_s(\cdot)$ work on each cluster independently. A cluster contains points in $\hat{\mathcal{P}}_u$ that is upsampled from the same point in decompressed sparse point cloud $\hat{\mathcal{P}}_s$ and the cluster centroid is set to this corresponding point in $\hat{\mathcal{P}}_s$. The workflow of the refinement layer involves the following two stages:
\begin{enumerate}
    \item In the compression phase, the relative positions between $r$ points within a cluster and the cluster centroid, along with the corresponding residuals of these $r$ points, are concatenated and then fed into the encoder $g_a(\cdot)$. For each cluster, the input is first mapped to features with two stacked ResNet blocks~\cite{he2016deep}, then aggregated with an attention-based weighting block~\cite{Boulch_2022_CVPR} and finally re-encoded into a latent feature vector. Subsequently, the latent features $y$ are quantized to $\hat{y}$ and entropy coded.
    \item In the decompression phase, the base layer reconstructs $\hat{\mathcal{P}}_u$ with the upsampling rate that is either aligned with the downsampling rate in the compression phase or flexibly configured with a desired number. Next, the INR-based decoder $g_s(\cdot)$ predicts residuals for each point in $\hat{\mathcal{P}}_u$ to yield a finer reconstruction $\hat{\mathcal{P}}$. To achieve this, for each query point in $\hat{\mathcal{P}}_u$, its associated latent feature is determined by using the row of $\hat{y}$ corresponding to the cluster centroid. Subsequently, the relative position between the query point and its corresponding cluster centroid, along with the associated latent feature, are fed into the decoder $g_s(\cdot)$ to estimate the residual. By adding the predicted residuals, points in $\hat{\mathcal{P}}_u$ are projected onto the learned implicit surface.
\end{enumerate}

\subsection{Graph-based conditional entropy model}
Unlike the common 1-D sequences with natural order such as audio or text, the latent features $\hat{y}$ generated by the encoder $g_a(\cdot)$ are arranged in an irregular and unordered layout. The lack of regularity and order poses challenges in forming local contexts, a crucial factor for coding efficiency. To our best knowledge, none of the existing point-based compression methods have effectively exploited local contexts for point cloud coding. The SOTA methods such as DPCC~\cite{he2022density} often rely on an unconditional fully factorized entropy model~\cite{balle2016end}, resulting in sub-optimal coding performance. 

In contrast to the existing methods, we propose a novel graph-based content-adaptive conditional entropy model that removes statistical redundancies among neighboring features against the irregularity of raw 3D point clouds.
Specifically, given the unordered latent features $\hat{y}$ (extracted from residuals) to be compressed, we first use KNN to characterize the neighboring relations of $\hat{y}$ based on the positions provided by the sparse point cloud $\hat{\mathcal{P}}_s$. As such, latent features $\hat{y}$ can be interpreted as a graph signal supported on a KNN graph with $m$ nodes where each node corresponds to a point in $\hat{\mathcal{P}}_s$. 
By employing graph structure, the CRCIR method can exploit the statistical correlation between latent features residing in neighboring nodes. This statistical dependency in a locality is captured by introducing a paired hyperencoder $h_a(\cdot)$ and hyperdecoder $h_s(\cdot)$ to learn a hyperprior~\cite{balle2018variational} on the latent representation $\hat{y}$.
Specifically, the probability models of $\hat{y}$ is assumed to be Gaussian distributions whose means and scales are conditioned on local content. The hyperencoder $h_a(\cdot)$ learns hyperprior $\hat{z}$ from latent features $\hat{y}$ and $\hat{\mathcal{P}}_s$. The hyperdecoder $h_s(\cdot)$ predicts the means and scales of $\hat{y}$ based on the learned hyperprior $\hat{z}$. 

\begin{figure}[!t]
    \centering
    \includegraphics[width=\linewidth]{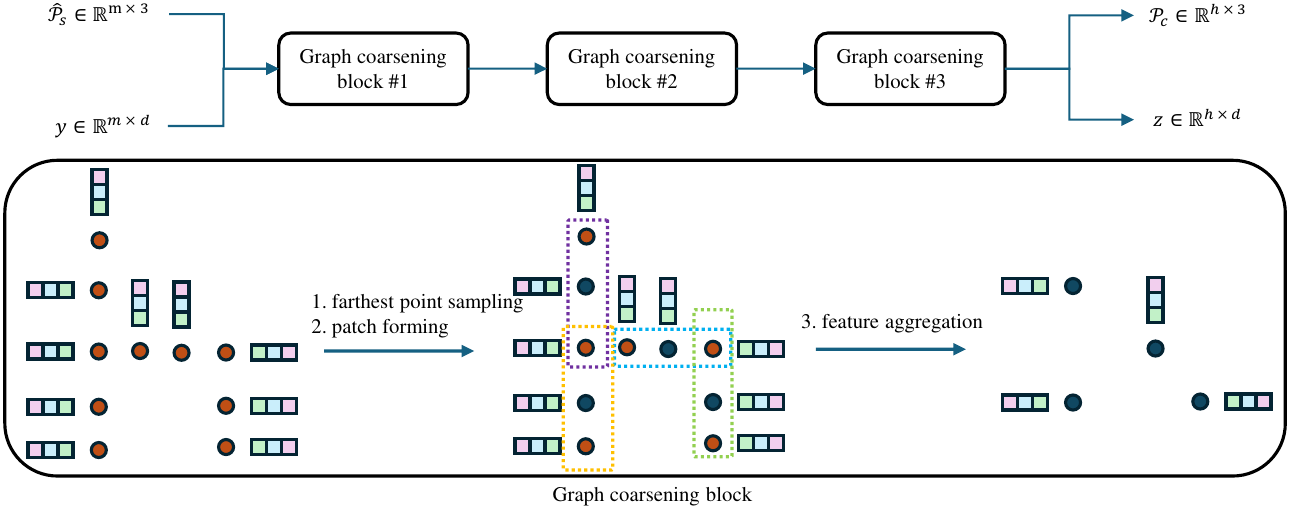}
    \caption{The architecture of the hyperencoder $h_a(\cdot)$. Specifically, the hyperencoder $h_a(\cdot)$ comprises three cascaded graph coarsening blocks, which extend the concept of downsampling kernels in regular structures to irregular graphs. The workflow of each block involves the following steps: 1. choosing seeds with FPS~\cite{qi2017pointnet++}; 2. forming the patch of K-nearest neighbors (KNN) for each seed; 3. extracting features for each patch. In step 3 features are aggregated with an EdgeConv layer whose edge function adopts the fifth option outline in~\cite{wang2019dynamic}. Additionally, the aggregation operation of each EdgeConv layer opts for the maximum pooling operator.}
    \label{fig:hyperencoder}
\end{figure}

\begin{figure}[!t]
    \centering
    \includegraphics[width=\linewidth]{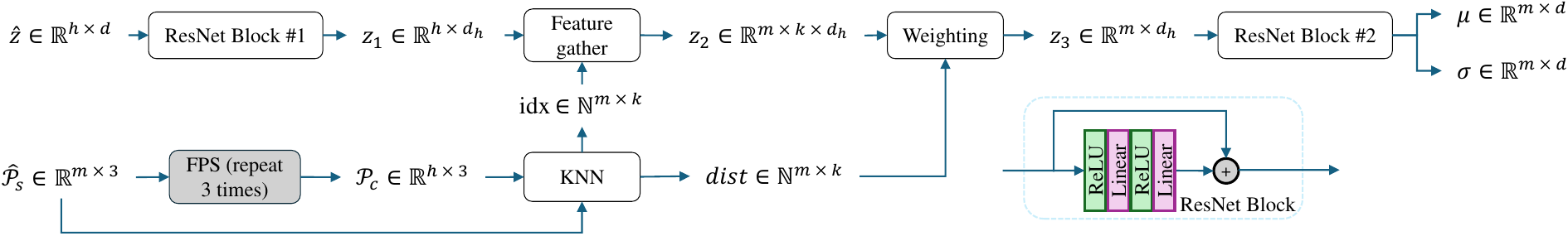}
    \caption{The workflow and architecture of the hyperdecoder $h_s(\cdot)$. }
    \label{fig:hyperdecoder}
\end{figure}

The detailed architectures of the proposed entropy model are as follows:
\begin{enumerate}
    \item As illustrated in \cref{fig:hyperencoder}, the hyperencoder $h_a(\cdot)$ progressively downsamples the graph-organized latent features $y$ to $z$ using stacked three graph coarsening blocks. 
    At the same time, $\hat{\mathcal{P}}_s$ is also downsampled to $\mathcal{P}_c$ with the same length of $z$.
    Each feature of $z$ corresponds to a distinct point in $\mathcal{P}_c$, while the features in $\hat{y}$ are aggregated and then quantized to $\hat{z}$ followed by being entropy coded with an unconditional factorized entropy model. Specifically, each row of $\hat{z}$ corresponds to a point in $\mathcal{P}_c$ and exhibits strong correlation with a set of neighboring features in $\hat{y}$. 
    \item As depicted in \cref{fig:hyperdecoder}, in the hyperdecoder $h_s(\cdot)$, $\hat{z}$ is re-encoded into $z_1$ and the positional information of $\hat{z}$ is derived by using three deterministic FPS processes to replicate $\hat{\mathcal{P}}_c$ where the downsampling rates match the values used in the compression phase. Next, for each point in $\hat{\mathcal{P}}_s$, its $k$ nearest neighbors in $\hat{\mathcal{P}}_c$ are characterized and the distances are recorded as $\{d_1,d_2,\cdots,d_k\}$ and $k$ features in $z_1$ are gathered. Subsequently, these $k$ features are combined using a weighted sum, where the weights are determined by $w_j=\frac{\exp(-\alpha d_j)}{\sum_{i=1}^{k}\exp(-\alpha d_i)},~j=1,\cdots,k$. Finally, the means and scales are predicted from $z_3$ with a ResNet block.
\end{enumerate}

\section{Experiments}
\subsection{Experimental setup}
\subsubsection{Training details} We implement the CRCIR method using PyTorch~\cite{paszke2019pytorch} and CompressAI~\cite{begaint2020compressai}, training it with three NVIDIA 1080 Ti GPUs. We use the Adam optimizer~\cite{kingma2014adam} by setting $\beta_1=0.9$ and $\beta_2=0.999$. The loss function is 
\begin{equation}
    \mathcal{L}=\ell_1(r,\hat{r})+\lambda R_{\hat{y}} + \lambda R_{\hat{z}}
    \label{eq:loss}
\end{equation}
where $\lambda$ is used to balance the rate-distortion trade-off, $R_{\hat{y}}$ and $R_{\hat{z}}$ are the rates of latent features and side information respectively. 
The training process involves two stages. Initially, only the distortion term in \cref{eq:loss} is minimized, with a learning rate of $10^{-2}$. In the subsequent stage, both the rate and distortion terms are jointly optimized, with a learning rate of $10^{-4}$. The first training stage comprises 34,410 steps, followed by the second stage of 10,519 steps, each with a batch size of 24.

\subsubsection{Baselines}We compare our method with two representative point-based compression methods: DPCC~\cite{he2022density} and 3QNet~\cite{huang20223qnet}. Besides, two SOTA non-learning-based methods are also included: MPEG G-PCC~\cite{graziosi2020overview} and Google Draco~\cite{galligan2018google}. Moreover, we place in the supplementrary material a detailed comparison with two sparse-tensor-based methods: PCGCv2~\cite{wang2021multiscale} and SparsePCGC~\cite{wang2022sparse}.

\subsubsection{Datasets}Our main experiments are carried out on the ShapeNet dataset~\cite{chang2015shapenet}. We follow~\cite{he2022density} to split training/testing/validation sets and sample points from meshes. The coordinates of all point clouds are normalized to $[-1, 1]$. For each shape, we densely sample its mesh representation at $120$k points to generate the point cloud. 
To evaluate the generalizability to unseen data, we also use the real-scanned point clouds in the scene-level Redwood indoor dataset~\cite{Park2017}. Furthermore, we provide experiments carried out on more diverse datasets~\cite{DBLP:conf/eccv/BerrettiBP12, dfaust:CVPR:2017,pandey2011ford} in the supplementary material.
\subsubsection{Metrics}
Following~\cite{he2022density}, we adopt point-to-point Chamfer distance (CD) and point-to-plane PSNR for geometry distortion and bits per point (bpp) for compression rate. When the ground truth mesh is available, we also use point-to-mesh distance (P2M) to evaluate the proximity to the ground truth surface. The above three distortion metrics are calculated using PyTorch3D~\cite{ravi2020pytorch3d} and we utilize the same method to calculate PSNR as used in~\cite{huang20223qnet}. When calculating these three metrics, we denormalize each point cloud to its original scale.

\begin{figure}[tb]
  \centering
  \begin{subfigure}{0.31\linewidth}
    \centering
    \includegraphics[width=1.6in]{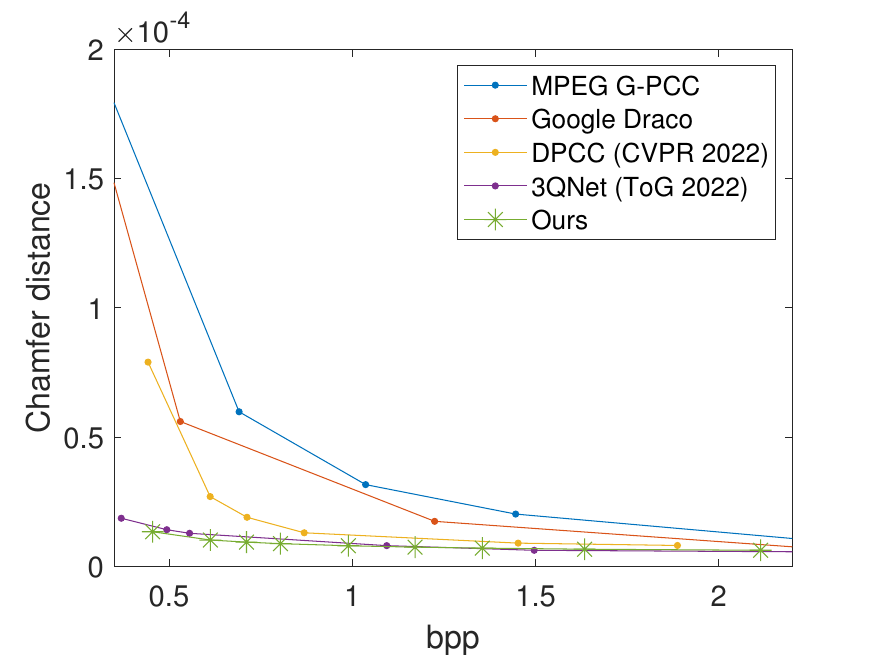}
    \caption{}
  \end{subfigure}
  \hfill
  \begin{subfigure}{0.31\linewidth}
    \centering
   \includegraphics[width=1.6in]{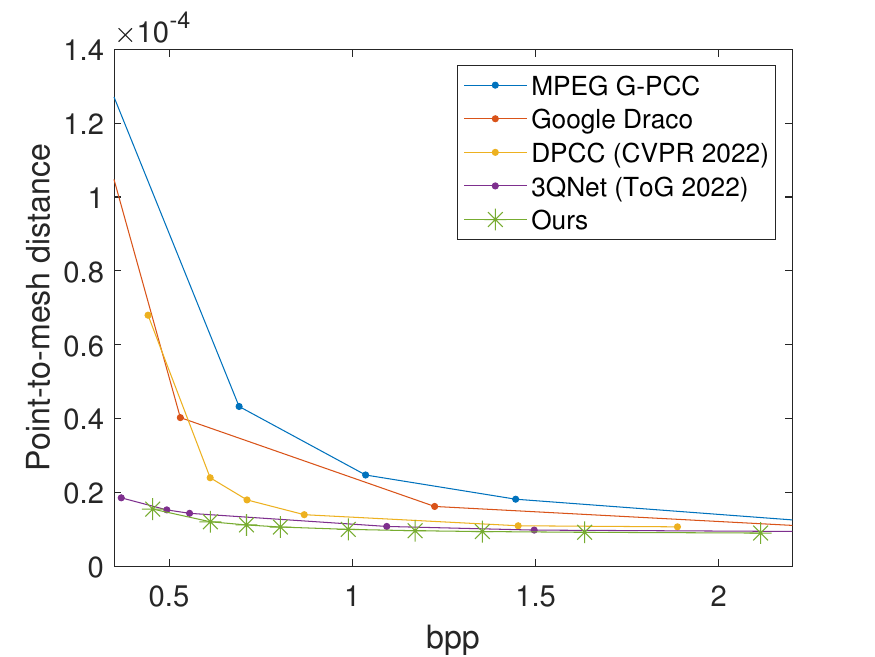}
   \caption{}
  \end{subfigure}
  \hfill
  \begin{subfigure}{0.31\linewidth}
   \centering
   \includegraphics[width=1.6in]{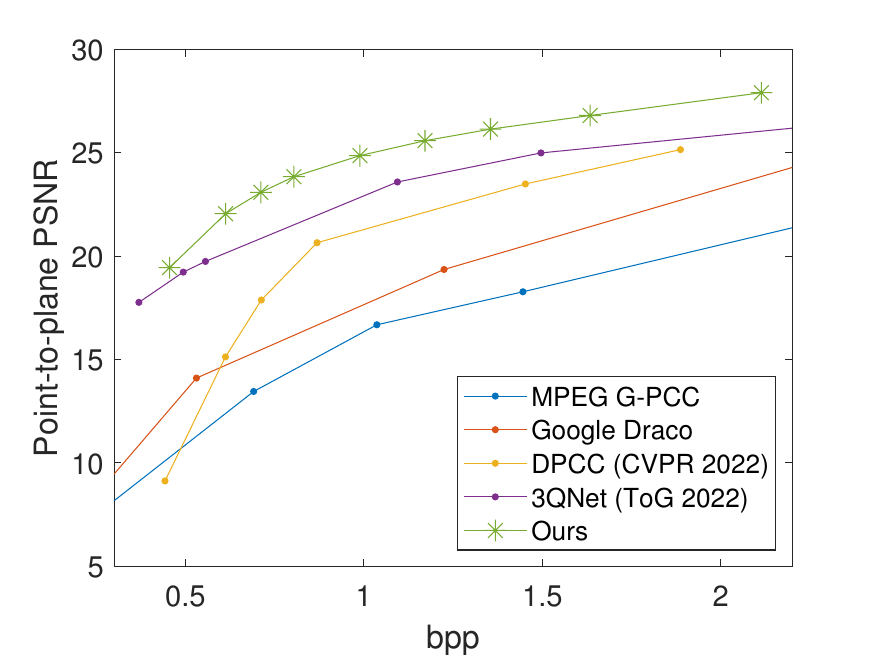}
   \caption{}
  \end{subfigure}
  \caption{Quantitative results on the ShapeNet dataset.}
  \label{fig:rd_curves}
\end{figure}
\subsection{Comparison with SOTA}
\subsubsection{Rate-distortion performance} 
In \cref{fig:rd_curves}, 
we present the RD-curves for three distortion metrics tested on the ShapeNet dataset. The CRCIR method achieves slight rate-distortion performance gain over 3QNet~\cite{huang20223qnet} in terms of the CD and P2M metrics, while we attain an improvement of approximately 2dB in the PSNR metric at the same bitrate (greater than 0.6 bpp).
Additionally, the CRCIR method significantly outperforms the other three baseline methods on all three metrics across the entire bitrate spectrum, with particularly remarkable performance gains at low bitrates. The inferiority of MPEG G-PCC~\cite{graziosi2020overview} and Google Draco~\cite{galligan2018google} at low bitrates mainly arises from their inability to handle raw point clouds directly. These two methods recude bitrate by increasing the quantization step and the resulting quantization error makes them perform poorly. Though DPCC ~\cite{he2022density} achieves direct processing of raw point clouds like ours, its failure to exploit local contexts for feature coding limits its ability to effectively leverage statistical dependencies among neighboring features. Consequently, it requires a higher bit allocation to accurately model local details, leading to poor performance at low bitrates. In comparison to all competing methods, the CRCIR method employ a graph-based conditional entropy model to compress features in the refinement layer. This entropy model adjusts to local geometry and exploits the correlation among neighboring features, leading to significant bitrate reduction. 
\subsubsection{Generalizability} 
\begin{figure}[tb]
  \centering
  \begin{subfigure}{0.45\linewidth}
    \centering
    \includegraphics[width=1.6in]{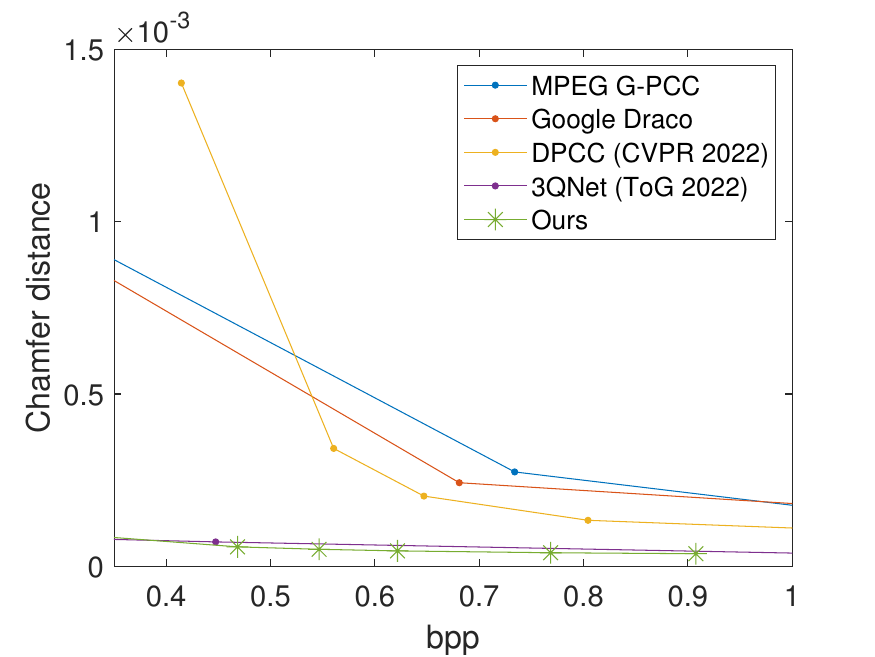}
    \caption{}
  \end{subfigure}
  \hfill
  \begin{subfigure}{0.45\linewidth}
    \centering
   \includegraphics[width=1.6in]{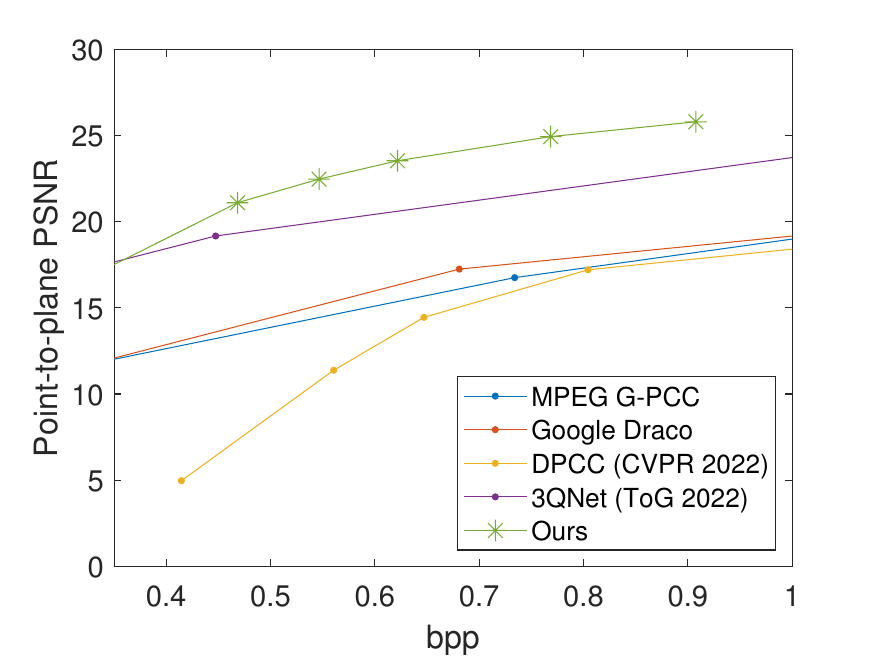}
   \caption{}
  \end{subfigure}
  \caption{Quantitative results on the scene-level Redwood indoor dataset.}
  \label{fig:rd_curve_livingroom}
\end{figure}

In \cref{fig:rd_curve_livingroom}, we present the performance on point clouds tested on the scene-level Redwood indoor dataset~\cite{Park2017}. 
Despite being trained on the ShapeNet dataset comprising man-made CAD objects only, the CRCIR method still demonstrates superiority in terms of rate-distortion performance on the real-scanned living rooms within this dataset. The experimental results highlight the generalizability of the CRCIR method to large-scale real-world data. 

\subsubsection{Error visualization} 
\begin{figure}
	\centering
        \includegraphics[width=\linewidth]{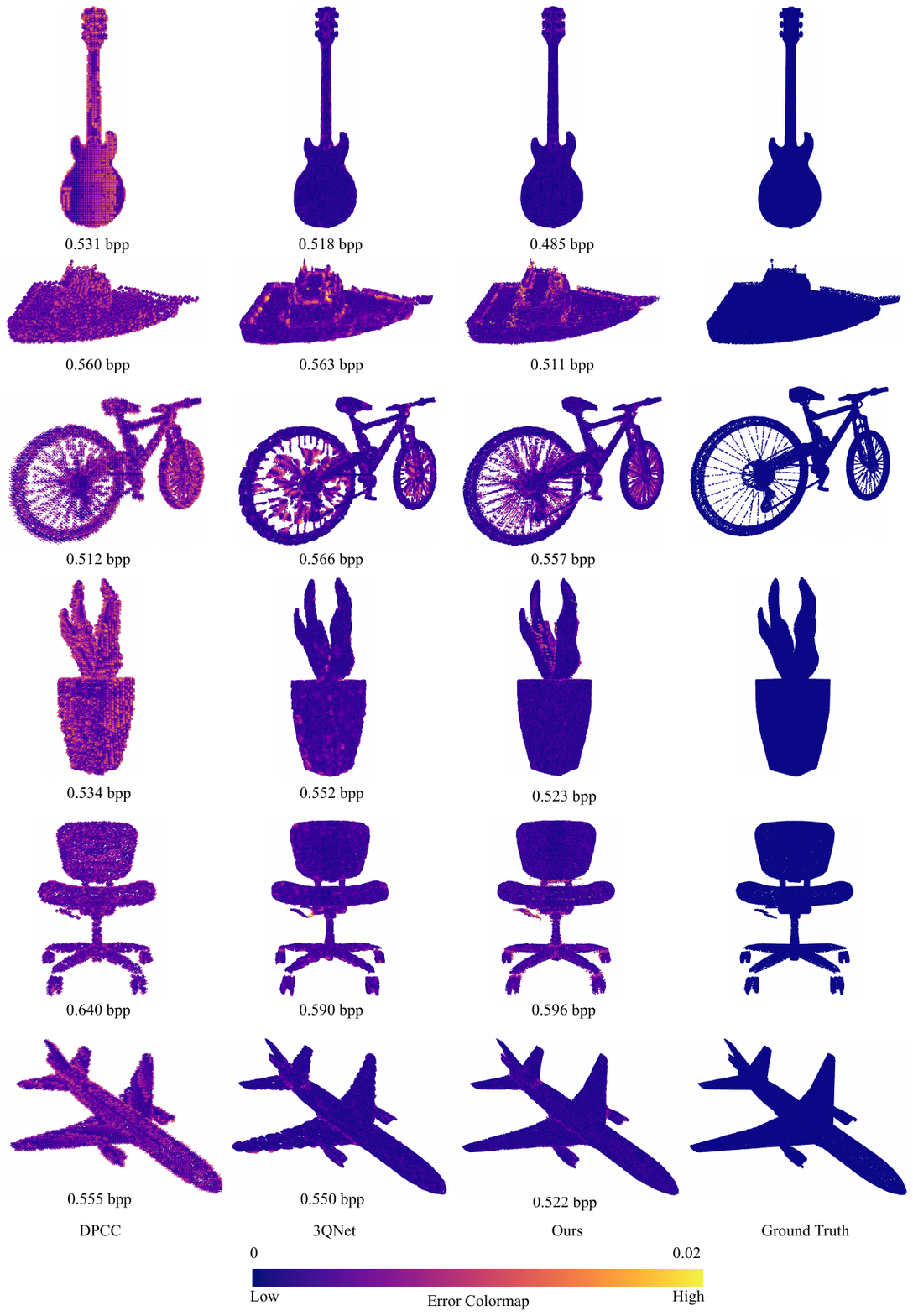}
	\caption{Perceptual comparison results on the ShapeNet dataset.}
	\label{fig:visuals}
\end{figure}

For a better perceptual understanding of the reconstruction qualities of the competing point-based methods, we visualize the reconstruction errors in \cref{fig:visuals}. Here we use the distance of a point to its nearest neighbors in the ground truth as the error metric, which is calculated with Open3D~\cite{Zhou2018}. One can appreciate the overall superb visual quality of the CRCIR method because of its ability to reconstruct point clouds with fewer errors, especially around areas containing finer detail like the bicycle spokes and the head stalk of the guitar.
\subsubsection{Complexity and latency} \cref{tab:efficiency_gpu} presents several key metrics related to both the memory and time complexities of the model, including the number of model parameters, memory footprint of executable files or checkpoints, GPU vRAM consumption, and per-object coding latency. The CRCIR method possesses the lowest model parameters and memory footprints. Regarding encoding and decoding runtime, the CRCIR method is only surpassed by Draco~\cite{galligan2018google}, but the increase in coding latency is negligible. 
In comparison to 3QNet~\cite{huang20223qnet}, which represents the SOTA approach, the CRCIR method achieves significant reductions in the number of model parameters, encoding latency, and decoding latency by factors of $381\times$, $158\times$ and $71\times$ respectively.  
Furthermore, the CRCIR method requires less GPU vRAM, making it compatible with most consumer-level GPU cards. These advantages are primarily attributed to the proposed dual-layer architecture. In this design, the non-learning base layer reconstructs main structures with very low complexity, while the refinement layer focuses on recovering details, enabling even small networks to perform effectively.
\begin{table}
	\caption{Comparison of different compression methods in model complexity, memory footprint and coding latency tested on the ShapeNet dataset.\label{tab:efficiency_gpu}
	}
	\centering
	\begin{tabular}{@{}lccccc@{}}
		\toprule
		\multirow{2}{*}{Methods}&Param.~$\downarrow$ & GPU vRAM~$\downarrow$& Memory~$\downarrow$&Enc. Time~$\downarrow$&Dec. Time~$\downarrow$\\
		&M&MB&MB&s&s\\
		\midrule
		GPCC~\cite{graziosi2020overview}&-&-&3.66&0.158&0.191\\
		Draco~\cite{galligan2018google}&-&-&2.72&\textbf{0.034}&\textbf{0.014}\\
		DPCC~\cite{he2022density}&0.088&858&0.44&1.271&0.395\\
		3QNet~\cite{huang20223qnet}&21.420&5050&81.71&6.867&1.447\\
		Ours&\textbf{0.056}&\textbf{854}&\textbf{0.26}&0.043&0.020\\
		\bottomrule
	\end{tabular}
\end{table}

\begin{table}[tb]
	\caption{Comparison of different methods in upsampling performance under various upsampling rates ranging from $\times2$ to $\times8$. DPCC, 3QNet, and our method consume 2.200, 2.135, and 2.426 bpp, respectively. \label{tab:superresolution}}
	\centering
	\begin{tabular}{lccclccclcccl}
		\toprule
		Methods&\multicolumn{3}{c}{DPCC~\cite{he2022density}+Grad-PU~\cite{he2023grad}}&&\multicolumn{3}{c}{3QNet~\cite{huang20223qnet}+Grad-PU~\cite{he2023grad}}&&\multicolumn{3}{c}{Ours}&\\ \cline{2-4}\cline{6-8}\cline{10-12}
		\multirow{2}{*}{Up rates}& CD~$\downarrow$    & P2M~$\downarrow$    & Time~$\downarrow$&&CD~$\downarrow$    & P2M~$\downarrow$    & Time~$\downarrow$ &           & CD~$\downarrow$    & P2M~$\downarrow$             & Time~$\downarrow$&    \\ 
		&$10^{-5}$&$10^{-5}$&s&&$10^{-5}$&$10^{-5}$&s&&$10^{-5}$&$10^{-5}$&s& \\ \midrule
		$\times2$&27.639&15.216&2.309&&26.622&19.033&2.415&&\textbf{10.922}&\textbf{5.238}&\textbf{0.172}&\\
		$\times3$&24.391&13.975&4.474&&23.441&18.016&4.589&&\textbf{7.699}&\textbf{4.756}&\textbf{0.173}&\\
		$\times4$&22.087&13.101&7.577&&21.572&17.450&7.818&&\textbf{6.262}&\textbf{4.614}&\textbf{0.171}&\\
		$\times5$&20.291&12.444&12.095&&20.354&17.086&12.195&&\textbf{5.678}&\textbf{4.632}&\textbf{0.170}&\\
		$\times6$&18.869&11.936&17.297&&19.439&16.783&17.484&&\textbf{5.396}&\textbf{4.695}&\textbf{0.173}&\\
		$\times7$&17.757&11.528&24.068&&18.715&16.501&24.295&&\textbf{5.243}&\textbf{4.780}&\textbf{0.173}&\\
		$\times8$&16.846&11.208&31.179&&18.164&16.276&31.230&&\textbf{5.171}&\textbf{4.873}&\textbf{0.170}&\\
		\bottomrule
	\end{tabular}
\end{table}
\subsection{Upsampling after decompression}
One intriguing feature of the proposed CRCIR method is its ability to function as an arbitrary-scale upsampling network.
To evaluate the upsampling performance, we compare it with two baselines: DPCC~\cite{he2022density}+Grad-PU~\cite{he2023grad} and 3QNet~\cite{huang20223qnet}+Grad-PU~\cite{he2023grad}, where Grad-PU is the SOTA arbitrary-scale upsampling method.
We conduct comparative evaluations on the PU-GAN dataset~\cite{li2019pu} using point clouds progressively downsampled from $120$k points to $15$k points. Specifically, each downsampling process removes $15$k points from the dense point cloud. The point cloud with $15$k points is compressed and the decompressed point cloud is fed into the upsampling network, which performs upsampling at rates ranging from $\times2$ to $\times8$. The remaining point clouds serve as the ground truth for corresponding upsampling rate. \cref{tab:superresolution} presents the distortion and the sum of decompression and upsampling runtime under different upsampling rates. The CRCIR method notably and consistently surpasses the competing baselines in both distortion and efficiency metrics. Besides, the runtime of our methods remains unchanged despite increasing the upsampling rate, whereas competing baselines suffer from a significant increase in runtime. 
This discrepancy in upsampling efficiency stems from the technical difference between Grad-PU and ours in the method to refine the initialized point cloud. Grad-PU requires iterative refinement and compute-intensive backward propagation, whereas the CRCIR method circumvent them by directly estimating the residual between the initialized point and the underlying surface. This substantially alleviate the increased workload associated with refining more initialized points.
\subsection{Ablation study}
\begin{figure}[tb]
  \centering
  \begin{subfigure}{0.31\linewidth}
    \centering
    \includegraphics[width=1.6in]{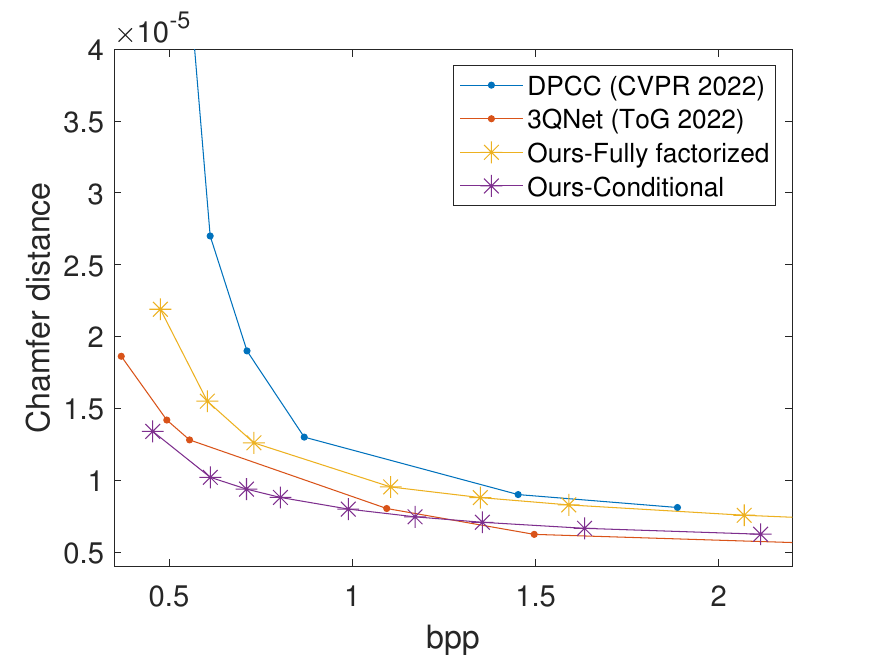}
    \caption{}
  \end{subfigure}
  \hfill
  \begin{subfigure}{0.31\linewidth}
    \centering
   \includegraphics[width=1.6in]{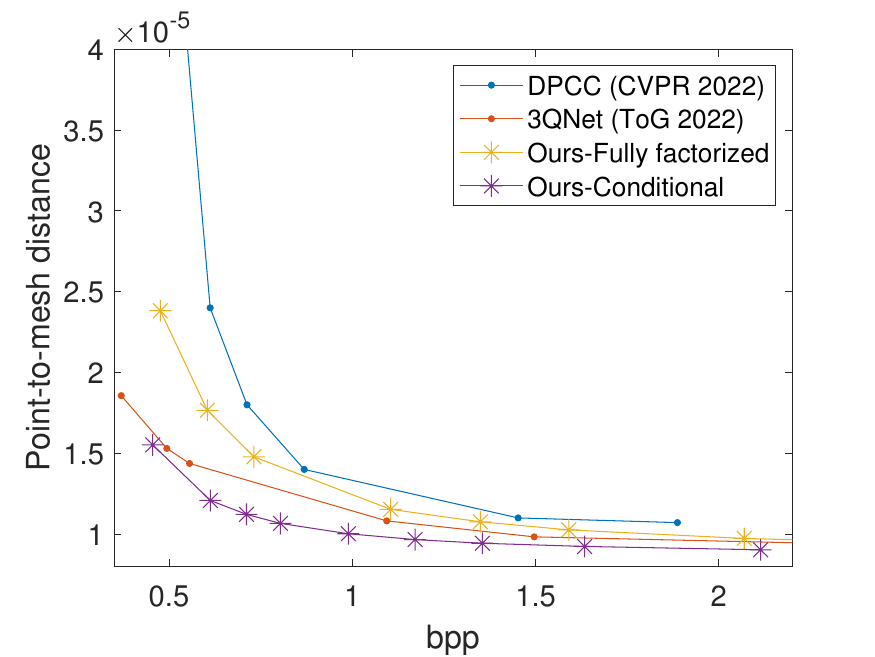}
   \caption{}
  \end{subfigure}
  \hfill
  \begin{subfigure}{0.31\linewidth}
   \centering
   \includegraphics[width=1.6in]{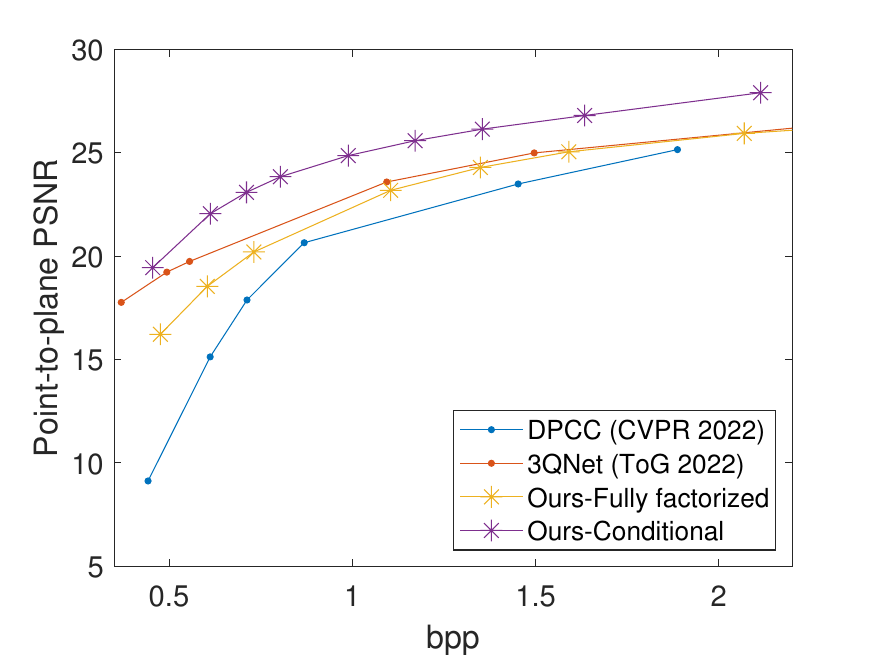}
   \caption{}
  \end{subfigure}
  \caption{Ablation study on the effectiveness of employing content-adaptive conditional entropy model in the refinement layer.}
  \label{fig:rd_curve_ablation}
\end{figure}

 In \cref{fig:rd_curve_ablation}, we present the effectiveness and necessity of exploiting content-aware conditional entropy model, using experimental results from the ShapeNet dataset. RD-curves of DPCC~\cite{he2022density} and 3QNet~\cite{huang20223qnet} are included for reference. Utilizing the proposed conditional entropy model drastically shifts R-D curves towards lower bitrates and enables our method to surpass 3QNet~\cite{huang20223qnet} whose complexity is two orders of magnitude higher. This implies that employing a conditional entropy model rather than a simple unconditional fully factorized entropy model is a key factor in achieving outstanding results.

\subsection{Limitation discussion}
By incorporating INR, we enable our decoder to sample points on the underlying surface at arbitrary densities. This allows our method to function as both a compression and arbitrary-scale upsampling network for 3D point clouds. However, when dealing with highly sparse input point clouds like LiDAR data, our method remains effective for compression but less so for arbitrary-scale upsampling. This is primarily because of the challenges involved in inferring plausible underlying surface from sparse and unevenly distributed LiDAR point clouds. To our knowledge, no existing INR-based upsampling methods exhibits remarkable capacity in effectively upsampling LiDAR point clouds at arbitrary scales.
\section{Conclusion}
We introduce a novel CRCIR method for compressing irregular point cloud geometry. It comprises a base layer tailored for reconstructing the main structures at low complexity and a refinement layer for recovering fine details. By employing the coarse geometry from base layer, we form local context in which the latent features in the refinement layer are compressed with a content-adaptive conditional entropy model. This allows the CRCIR method to remove the redundancies among neighboring features, leading to significant bitrate saving. By incorporating INR into the refinement layer, the CRCIR method possesses the respective advantages of NTC and INR in streamlining the compression process and sampling points on the learned implicit surface at arbitrary densities. Experimental results demonstrate the superiority of our method over existing methods, achieving outstanding rate-distortion performance, the remarkable ability to work as an arbitrary-scale upampling network and significant reductions in model complexity and coding latency.

\section*{Acknowledgements}
This work was supported by the Natural Sciences and Engineering Research Council of Canada.
%
%
\bibliographystyle{splncs04}
\bibliography{main}

\begin{thebibliography}{10}
\providecommand{\url}[1]{\texttt{#1}}
\providecommand{\urlprefix}{URL }
\providecommand{\doi}[1]{https://doi.org/#1}

\bibitem{balle2016end}
Ball{\'e}, J., Laparra, V., Simoncelli, E.P.: End-to-end optimized image compression. arXiv preprint arXiv:1611.01704  (2016)

\bibitem{balle2018variational}
Ball{\'e}, J., Minnen, D., Singh, S., Hwang, S.J., Johnston, N.: Variational image compression with a scale hyperprior. arXiv preprint arXiv:1802.01436  (2018)

\bibitem{begaint2020compressai}
B{\'e}gaint, J., Racap{\'e}, F., Feltman, S., Pushparaja, A.: Compressai: a pytorch library and evaluation platform for end-to-end compression research. arXiv preprint arXiv:2011.03029  (2020)

\bibitem{DBLP:conf/eccv/BerrettiBP12}
Berretti, S., Bimbo, A.D., Pala, P.: Superfaces: A super-resolution model for 3d faces. In: ECCV Workshops (1). pp. 73--82 (2012)

\bibitem{dfaust:CVPR:2017}
Bogo, F., Romero, J., Pons-Moll, G., Black, M.J.: Dynamic {FAUST}: {R}egistering human bodies in motion. In: IEEE Conf. on Computer Vision and Pattern Recognition (CVPR) (Jul 2017)

\bibitem{Boulch_2022_CVPR}
Boulch, A., Marlet, R.: Poco: Point convolution for surface reconstruction. In: Proceedings of the IEEE/CVF Conference on Computer Vision and Pattern Recognition (CVPR). pp. 6302--6314 (June 2022)

\bibitem{chang2015shapenet}
Chang, A.X., Funkhouser, T., Guibas, L., Hanrahan, P., Huang, Q., Li, Z., Savarese, S., Savva, M., Song, S., Su, H., et~al.: Shapenet: An information-rich 3d model repository. arXiv preprint arXiv:1512.03012  (2015)

\bibitem{cheng2020learned}
Cheng, Z., Sun, H., Takeuchi, M., Katto, J.: Learned image compression with discretized gaussian mixture likelihoods and attention modules. In: Proceedings of the IEEE/CVF conference on computer vision and pattern recognition. pp. 7939--7948 (2020)

\bibitem{choy20194d}
Choy, C., Gwak, J., Savarese, S.: 4d spatio-temporal convnets: Minkowski convolutional neural networks. In: Proceedings of the IEEE Conference on Computer Vision and Pattern Recognition. pp. 3075--3084 (2019)

\bibitem{dong2015image}
Dong, C., Loy, C.C., He, K., Tang, X.: Image super-resolution using deep convolutional networks. IEEE transactions on pattern analysis and machine intelligence  \textbf{38}(2),  295--307 (2015)

\bibitem{dupont2021coin}
Dupont, E., Goli{\'n}ski, A., Alizadeh, M., Teh, Y.W., Doucet, A.: Coin: Compression with implicit neural representations. arXiv preprint arXiv:2103.03123  (2021)

\bibitem{dupont2022coin++}
Dupont, E., Loya, H., Alizadeh, M., Goli{\'n}ski, A., Teh, Y.W., Doucet, A.: Coin++: Neural compression across modalities. arXiv preprint arXiv:2201.12904  (2022)

\bibitem{feng2022neural}
Feng, W., Li, J., Cai, H., Luo, X., Zhang, J.: Neural points: Point cloud representation with neural fields for arbitrary upsampling. In: Proceedings of the IEEE/CVF Conference on Computer Vision and Pattern Recognition. pp. 18633--18642 (2022)

\bibitem{fu2022octattention}
Fu, C., Li, G., Song, R., Gao, W., Liu, S.: Octattention: Octree-based large-scale contexts model for point cloud compression. In: Proceedings of the AAAI Conference on Artificial Intelligence. vol.~36, pp. 625--633 (2022)

\bibitem{galligan2018google}
Galligan, F., Hemmer, M., Stava, O., Zhang, F., Brettle, J.: Google/draco: a library for compressing and decompressing 3d geometric meshes and point clouds (2018)

\bibitem{graziosi2020overview}
Graziosi, D., Nakagami, O., Kuma, S., Zaghetto, A., Suzuki, T., Tabatabai, A.: An overview of ongoing point cloud compression standardization activities: Video-based (v-pcc) and geometry-based (g-pcc). APSIPA Transactions on Signal and Information Processing  \textbf{9}, ~e13 (2020)

\bibitem{guo2022data}
Guo, Y., Wu, X., Shu, X.: Data acquisition and preparation for dual-reference deep learning of image super-resolution. IEEE Transactions on Image Processing  \textbf{31},  4393--4404 (2022)

\bibitem{guo2020deep}
Guo, Y., Zhang, X., Wu, X.: Deep multi-modality soft-decoding of very low bit-rate face videos. In: Proceedings of the 28th ACM International Conference on Multimedia. pp. 3947--3955 (2020)

\bibitem{he2022elic}
He, D., Yang, Z., Peng, W., Ma, R., Qin, H., Wang, Y.: Elic: Efficient learned image compression with unevenly grouped space-channel contextual adaptive coding. In: Proceedings of the IEEE/CVF Conference on Computer Vision and Pattern Recognition. pp. 5718--5727 (2022)

\bibitem{he2016deep}
He, K., Zhang, X., Ren, S., Sun, J.: Deep residual learning for image recognition. In: Proceedings of the IEEE conference on computer vision and pattern recognition. pp. 770--778 (2016)

\bibitem{he2022density}
He, Y., Ren, X., Tang, D., Zhang, Y., Xue, X., Fu, Y.: Density-preserving deep point cloud compression. In: Proceedings of the IEEE/CVF Conference on Computer Vision and Pattern Recognition. pp. 2333--2342 (2022)

\bibitem{he2023grad}
He, Y., Tang, D., Zhang, Y., Xue, X., Fu, Y.: Grad-pu: Arbitrary-scale point cloud upsampling via gradient descent with learned distance functions. In: Proceedings of the IEEE/CVF Conference on Computer Vision and Pattern Recognition. pp. 5354--5363 (2023)

\bibitem{hu2019meta}
Hu, X., Mu, H., Zhang, X., Wang, Z., Tan, T., Sun, J.: Meta-sr: A magnification-arbitrary network for super-resolution. In: Proceedings of the IEEE/CVF conference on computer vision and pattern recognition. pp. 1575--1584 (2019)

\bibitem{huang2020octsqueeze}
Huang, L., Wang, S., Wong, K., Liu, J., Urtasun, R.: Octsqueeze: Octree-structured entropy model for lidar compression. In: Proceedings of the IEEE/CVF conference on computer vision and pattern recognition. pp. 1313--1323 (2020)

\bibitem{huang20193d}
Huang, T., Liu, Y.: 3d point cloud geometry compression on deep learning. In: Proceedings of the 27th ACM international conference on multimedia. pp. 890--898 (2019)

\bibitem{huang20223qnet}
Huang, T., Zhang, J., Chen, J., Ding, Z., Tai, Y., Zhang, Z., Wang, C., Liu, Y.: 3qnet: 3d point cloud geometry quantization compression network. ACM Transactions on Graphics (TOG)  \textbf{41}(6),  1--13 (2022)

\bibitem{ioffe2015batch}
Ioffe, S., Szegedy, C.: Batch normalization: Accelerating deep network training by reducing internal covariate shift. In: International conference on machine learning. pp. 448--456. pmlr (2015)

\bibitem{kim2022joint}
Kim, J.H., Heo, B., Lee, J.S.: Joint global and local hierarchical priors for learned image compression. In: Proceedings of the IEEE/CVF Conference on Computer Vision and Pattern Recognition. pp. 5992--6001 (2022)

\bibitem{kingma2014adam}
Kingma, D.P., Ba, J.: Adam: A method for stochastic optimization. arXiv preprint arXiv:1412.6980  (2014)

\bibitem{li2019pu}
Li, R., Li, X., Fu, C.W., Cohen-Or, D., Heng, P.A.: Pu-gan: a point cloud upsampling adversarial network. In: Proceedings of the IEEE/CVF international conference on computer vision. pp. 7203--7212 (2019)

\bibitem{li2021point}
Li, R., Li, X., Heng, P.A., Fu, C.W.: Point cloud upsampling via disentangled refinement. In: Proceedings of the IEEE/CVF conference on computer vision and pattern recognition. pp. 344--353 (2021)

\bibitem{liu2023learned}
Liu, J., Sun, H., Katto, J.: Learned image compression with mixed transformer-cnn architectures. In: Proceedings of the IEEE/CVF conference on computer vision and pattern recognition. pp. 14388--14397 (2023)

\bibitem{luo2021functional}
Luo, F., Wu, X., Guo, Y.: Functional neural networks for parametric image restoration problems. Advances in Neural Information Processing Systems  \textbf{34},  6762--6775 (2021)

\bibitem{luo2024and}
Luo, F., Wu, X., Guo, Y.: And: Adversarial neural degradation for learning blind image super-resolution. Advances in Neural Information Processing Systems  \textbf{36} (2024)

\bibitem{minnen2018joint}
Minnen, D., Ball{\'e}, J., Toderici, G.D.: Joint autoregressive and hierarchical priors for learned image compression. Advances in neural information processing systems  \textbf{31} (2018)

\bibitem{pandey2011ford}
Pandey, G., McBride, J.R., Eustice, R.M.: Ford campus vision and lidar data set. The International Journal of Robotics Research  \textbf{30}(13),  1543--1552 (2011)

\bibitem{pang2022grasp}
Pang, J., Lodhi, M.A., Tian, D.: Grasp-net: Geometric residual analysis and synthesis for point cloud compression. In: Proceedings of the 1st International Workshop on Advances in Point Cloud Compression, Processing and Analysis. pp. 11--19 (2022)

\bibitem{Park2017}
Park, J., Zhou, Q.Y., Koltun, V.: Colored point cloud registration revisited. In: ICCV (2017)

\bibitem{paszke2019pytorch}
Paszke, A., Gross, S., Massa, F., Lerer, A., Bradbury, J., Chanan, G., Killeen, T., Lin, Z., Gimelshein, N., Antiga, L., et~al.: Pytorch: An imperative style, high-performance deep learning library. Advances in neural information processing systems  \textbf{32} (2019)

\bibitem{peng2020convolutional}
Peng, S., Niemeyer, M., Mescheder, L., Pollefeys, M., Geiger, A.: Convolutional occupancy networks. In: Computer Vision--ECCV 2020: 16th European Conference, Glasgow, UK, August 23--28, 2020, Proceedings, Part III 16. pp. 523--540. Springer (2020)

\bibitem{postels20233d}
Postels, J., Str{\"u}mpler, Y., Reichard, K., Van~Gool, L., Tombari, F.: 3d compression using neural fields. arXiv preprint arXiv:2311.13009  (2023)

\bibitem{qi2017pointnet++}
Qi, C.R., Yi, L., Su, H., Guibas, L.J.: Pointnet++: Deep hierarchical feature learning on point sets in a metric space. Advances in neural information processing systems  \textbf{30} (2017)

\bibitem{qian2021pu}
Qian, G., Abualshour, A., Li, G., Thabet, A., Ghanem, B.: Pu-gcn: Point cloud upsampling using graph convolutional networks. In: Proceedings of the IEEE/CVF Conference on Computer Vision and Pattern Recognition. pp. 11683--11692 (2021)

\bibitem{qian2021deep}
Qian, Y., Hou, J., Kwong, S., He, Y.: Deep magnification-flexible upsampling over 3d point clouds. IEEE Transactions on Image Processing  \textbf{30},  8354--8367 (2021)

\bibitem{que2021voxelcontext}
Que, Z., Lu, G., Xu, D.: Voxelcontext-net: An octree based framework for point cloud compression. In: Proceedings of the IEEE/CVF Conference on Computer Vision and Pattern Recognition. pp. 6042--6051 (2021)

\bibitem{ravi2020pytorch3d}
Ravi, N., Reizenstein, J., Novotny, D., Gordon, T., Lo, W.Y., Johnson, J., Gkioxari, G.: Accelerating 3d deep learning with pytorch3d. arXiv:2007.08501  (2020)

\bibitem{saharia2022image}
Saharia, C., Ho, J., Chan, W., Salimans, T., Fleet, D.J., Norouzi, M.: Image super-resolution via iterative refinement. IEEE transactions on pattern analysis and machine intelligence  \textbf{45}(4),  4713--4726 (2022)

\bibitem{shi2016real}
Shi, W., Caballero, J., Husz{\'a}r, F., Totz, J., Aitken, A.P., Bishop, R., Rueckert, D., Wang, Z.: Real-time single image and video super-resolution using an efficient sub-pixel convolutional neural network. In: Proceedings of the IEEE conference on computer vision and pattern recognition. pp. 1874--1883 (2016)

\bibitem{song2023efficient}
Song, R., Fu, C., Liu, S., Li, G.: Efficient hierarchical entropy model for learned point cloud compression. In: Proceedings of the IEEE/CVF Conference on Computer Vision and Pattern Recognition. pp. 14368--14377 (2023)

\bibitem{strumpler2022implicit}
Str{\"u}mpler, Y., Postels, J., Yang, R., Gool, L.V., Tombari, F.: Implicit neural representations for image compression. In: European Conference on Computer Vision. pp. 74--91. Springer (2022)

\bibitem{wang2022sparse}
Wang, J., Ding, D., Li, Z., Feng, X., Cao, C., Ma, Z.: Sparse tensor-based multiscale representation for point cloud geometry compression. IEEE Transactions on Pattern Analysis and Machine Intelligence  (2022)

\bibitem{wang2021multiscale}
Wang, J., Ding, D., Li, Z., Ma, Z.: Multiscale point cloud geometry compression. In: 2021 Data Compression Conference (DCC). pp. 73--82. IEEE (2021)

\bibitem{wang2021lossy}
Wang, J., Zhu, H., Liu, H., Ma, Z.: Lossy point cloud geometry compression via end-to-end learning. IEEE Transactions on Circuits and Systems for Video Technology  \textbf{31}(12),  4909--4923 (2021)

\bibitem{wang2019dynamic}
Wang, Y., Sun, Y., Liu, Z., Sarma, S.E., Bronstein, M.M., Solomon, J.M.: Dynamic graph cnn for learning on point clouds. ACM Transactions on Graphics (tog)  \textbf{38}(5),  1--12 (2019)

\bibitem{wiesmann2021deep}
Wiesmann, L., Milioto, A., Chen, X., Stachniss, C., Behley, J.: Deep compression for dense point cloud maps. IEEE Robotics and Automation Letters  \textbf{6}(2),  2060--2067 (2021)

\bibitem{yan2019deep}
Yan, W., Liu, S., Li, T.H., Li, Z., Li, G., et~al.: Deep autoencoder-based lossy geometry compression for point clouds. arXiv preprint arXiv:1905.03691  (2019)

\bibitem{ye2021meta}
Ye, S., Chen, D., Han, S., Wan, Z., Liao, J.: Meta-pu: An arbitrary-scale upsampling network for point cloud. IEEE transactions on visualization and computer graphics  \textbf{28}(9),  3206--3218 (2021)

\bibitem{yifan2019patch}
Yifan, W., Wu, S., Huang, H., Cohen-Or, D., Sorkine-Hornung, O.: Patch-based progressive 3d point set upsampling. In: Proceedings of the IEEE/CVF Conference on Computer Vision and Pattern Recognition. pp. 5958--5967 (2019)

\bibitem{you2021patch}
You, K., Gao, P.: Patch-based deep autoencoder for point cloud geometry compression. In: ACM Multimedia Asia, pp.~1--7 (2021)

\bibitem{you2022ipdae}
You, K., Gao, P., Li, Q.: Ipdae: Improved patch-based deep autoencoder for lossy point cloud geometry compression. In: Proceedings of the 1st International Workshop on Advances in Point Cloud Compression, Processing and Analysis. pp. 1--10 (2022)

\bibitem{yu2018pu}
Yu, L., Li, X., Fu, C.W., Cohen-Or, D., Heng, P.A.: Pu-net: Point cloud upsampling network. In: Proceedings of the IEEE conference on computer vision and pattern recognition. pp. 2790--2799 (2018)

\bibitem{zhang2023yoga}
Zhang, J., Chen, T., Ding, D., Ma, Z.: Yoga: Yet another geometry-based point cloud compressor. In: Proceedings of the 31st ACM International Conference on Multimedia. pp. 9070--9081 (2023)

\bibitem{zhang2021attention}
Zhang, X., Wu, X.: Attention-guided image compression by deep reconstruction of compressive sensed saliency skeleton. In: Proceedings of the IEEE/CVF Conference on Computer Vision and Pattern Recognition. pp. 13354--13364 (2021)

\bibitem{zhang2023lvqac}
Zhang, X., Wu, X.: Lvqac: Lattice vector quantization coupled with spatially adaptive companding for efficient learned image compression. In: Proceedings of the IEEE/CVF Conference on Computer Vision and Pattern Recognition. pp. 10239--10248 (2023)

\bibitem{zhao2022self}
Zhao, W., Liu, X., Zhong, Z., Jiang, J., Gao, W., Li, G., Ji, X.: Self-supervised arbitrary-scale point clouds upsampling via implicit neural representation. In: Proceedings of the IEEE/CVF Conference on Computer Vision and Pattern Recognition. pp. 1999--2007 (2022)

\bibitem{Zhou2018}
Zhou, Q.Y., Park, J., Koltun, V.: {Open3D}: {A} modern library for {3D} data processing. arXiv:1801.09847  (2018)

\end{thebibliography}

\clearpage

\title{Supplementary Material of 'Fast Point Cloud Geometry Compression with Context-based Residual Coding and INR-based Refinement'} 

\titlerunning{Supplementary material for point cloud geometry compression with CRCIR}

\author{Hao Xu\inst{1}\orcidlink{0000-0001-5685-5225} \and
Xi Zhang\inst{2}\orcidlink{0000-0002-1993-6031} \and
Xiaolin Wu\inst{1}\thanks{Corresponding author.}\orcidlink{0000-0002-0103-5374}}

\authorrunning{H. Xu et al.}

\institute{McMaster University \and
Shanghai Jiao Tong University
\email{xu338@mcmaster.ca,~xzhang9308@gmail.com,~xwu@ece.mcmaster.ca}}

\maketitle
    \section{Additional implementation details}
The base layer employs Google Draco~\cite{galligan2018google} to compress the downsampled point cloud $\mathcal{P}_s$. For this compression, the quantization parameter is set to 9, and the compression level is set to 10. 

 The decoder $g_s(\cdot)$ in our compression system has a similar architecture to the decoder in~\cite{peng2020convolutional}; the differences are in the number and dimension of hidden layers (3 and 64 respectively in our case). The detailed hyperparameters of other network components are tabulated in \cref{tab:hyperpara_up}.

 Following 3QNet~\cite{huang20223qnet}, we utilize a single model to achieve variable bitrate compression by adjusting the size of the downsampled point cloud $\mathcal{P}_s$. This model is trained with the rate-distortion trade-off parameter, $\lambda$, set to $5 \times 10^{-3}$. 
 The selected downsampling rate ranges from $\times48$ to $\times6$, specifically including $$\{\times48,\times30,\times24,\times20,\times15,\times12,\times10,\times8,\times6\}.$$
 Note that these downsampling rates are examples; the actual sizes for $\mathcal{P}_s$ can be adjusted according to user needs and are not strictly limited to the values listed.
 During training, the size of $\mathcal{P}_s$ is set to $4000$ points, corresponding to a downsampling rate of $\times30$. Experimental results show that the model, trained at this specific downsampling rate, generalizes effectively to various other downsampling rates. 
        \begin{table}[h]
    \caption{The hyperparameters of convolutional layers where (i, o, k) denotes the number of channels in the input and output, and the kernel size respectively, and BN is the abbreviation of batch normalization~\cite{ioffe2015batch}.}
    \centering
	\begin{tabular}{ccccccc}
		\toprule
		Component&Layer&i&o&k&Normalization&Activation\\ \midrule
		\multirow{5}{*}{Encoder $g_a(\cdot)$}&ResNet block \#1&6&64&1&BN&ReLU\\
		&ResNet block \#2&64&64&1&BN&ReLU\\ 
		&Conv (upper)&64&64&1&-&-\\
		&Conv (lower)&64&1&1&-&-\\
		&ResNet block \#3&64&8&1&BN&ReLU\\
            \hline
            \multirow{3}{*}{Hyperencoder $h_a(\cdot)$}&Graph coarsening block \#1&8&32&1&-&ReLU\\
            &Graph coarsening block \#2&32&32&1&-&ReLU\\
            &Graph coarsening block \#3&32&8&1&-&ReLU\\
            \hline
            \multirow{2}{*}{Hyperdecoder $h_s(\cdot)$}&ResNet block \#1&8&32&1&-&ReLU\\
            &ResNet block \#2&32&16&1&-&ReLU\\
		\bottomrule
	\end{tabular}
	
	\label{tab:hyperpara_up}
    \end{table}	

\section{The selection of interpolation weights}
	Our non-learning base layer employs a simple but effective upsampling method to predict a dense point cloud $\hat{\mathcal{P}}_u$ from the downsampled and decompressed point cloud $\hat{\mathcal{P}}_s$, 
	which is defined as: 
	\begin{equation}
	    \hat{p}_u^{(ij)}=\hat{p}_s^{(i)}+u_{ij}(\hat{p}_s^{(ij)}-\hat{p}_s^{(i)}),~1\leq i\leq m,~1\leq j\leq r.
	    \label{eq:interp}
	\end{equation}
	Here $\hat{p}_s^{(i)}$ represents the $i$-th point in $\hat{\mathcal{P}}_s$, the set $\{\hat{p}_s^{(ij)}|1\leq j\leq r\}$ comprises the nearest $r$ neighbors of $\hat{p}_s^{(i)}$ in $\hat{\mathcal{P}}_s$, and $u_{ij}$ denotes the associated interpolation weights. 
    When interpolating a point between $\hat{p}_s^{(i)}$ and its $j$-th nearest neighbor $\hat{p}_s^{(ij)}$, the objective is to determine the weight $u_{ij}$ that minimizes the offset between this interpolated point and the underlying surface, ideally approaching $\mathop{0}\limits ^{\rightarrow}$. The golden-section search method is a fundamental tool for such optimization problems. 
    Inspired by this, we set the interpolation weights using the golden ratio. Specifically, we set $u_{ij}$ to $1-\frac{\sqrt{5}-1}{2}$. In this context, this approach can be seen as using the points predicted by a one-step golden section search as the upsampled point cloud. To mitigate significant deviations of interpolated points from $\hat{p}_s^{(i)}$, we penalize the weights for those distant neighbors. Hence, for $j>8$, the weight $u_{ij}$ is set to $0.75\times(1-\frac{\sqrt{5}-1}{2})$. Fine-tuning these hyperparameters meticulously is unnecessary, as the interpolation process aims solely to provide a satisfactory initialization for the subsequent learned refinement layer. This refinement layer will then learn how to improve the quality of the upsampled point cloud. We will address learning the optimal interpolation weights in future work.
 \section{Comparison with sparse-tensor-based methods}
\begin{figure}
		\centering
		\includegraphics[width=0.9\linewidth]{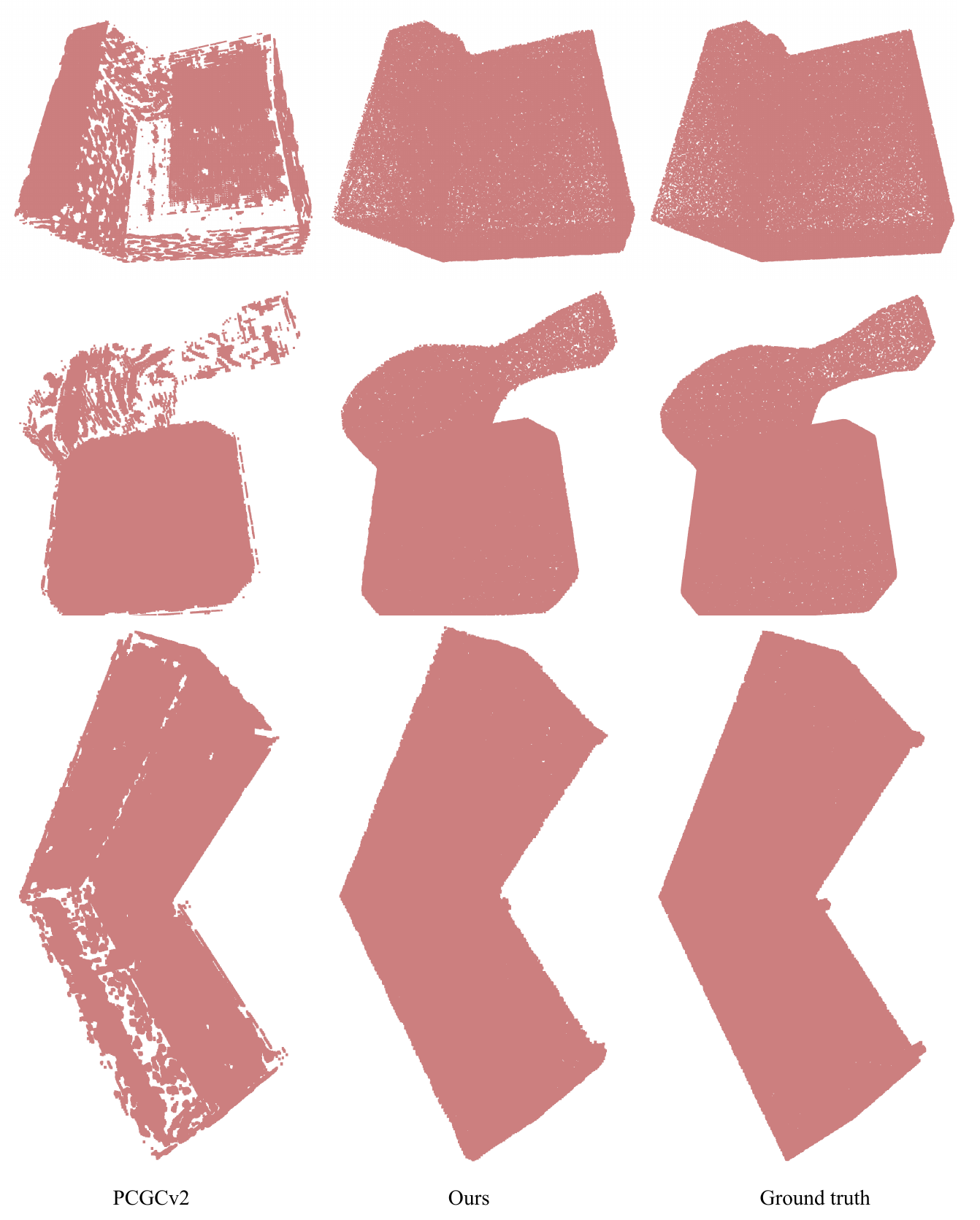}
		\caption{Perceptual comparisons with sparse-tensor-based methods.}
		\label{fig:missing}
	\end{figure}

	The sparse-tensor-based methods such as PCGCv2~\cite{wang2021multiscale} and SparsePCGC~\cite{wang2022sparse} use a memory-efficient data structure called sparse tensor~\cite{choy20194d} to store the voxelized geometry. By introducing regularity with such regular structure, the compression system can form local contexts for conditional entropy coding. However, due to quantization, these methods often suffer from severe artifacts, such as missing points in their reconstructions. In \cref{fig:missing}, we demonstrate this failure of the sparse-tensor-based approach. In contrast, our method can well preserve the original density information through direct processing of raw point clouds.
	
\begin{figure}
		\centering
		\subfloat[]{\includegraphics[width=1.6in]{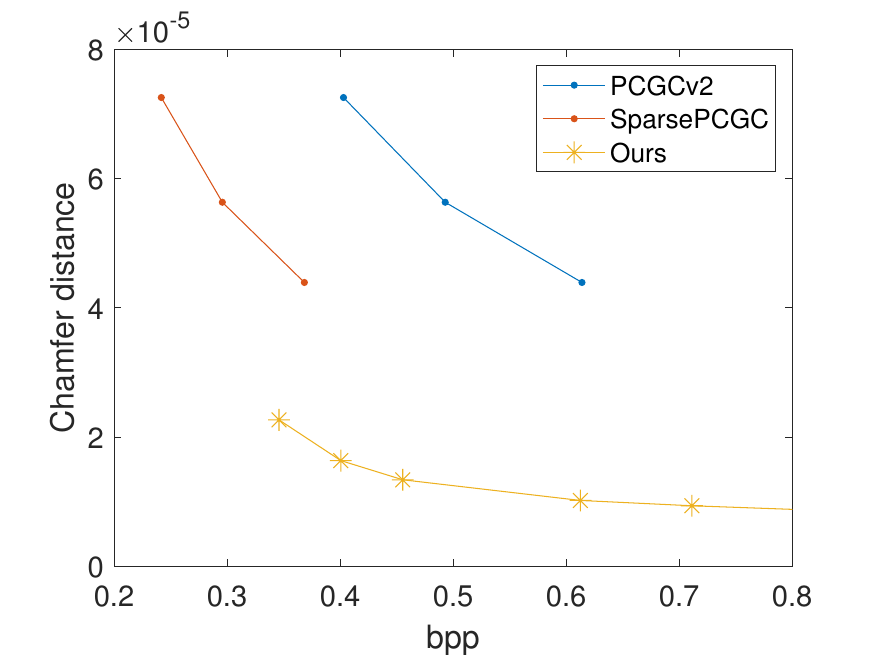}}%
	\hfil
	\subfloat[]{\includegraphics[width=1.6in]{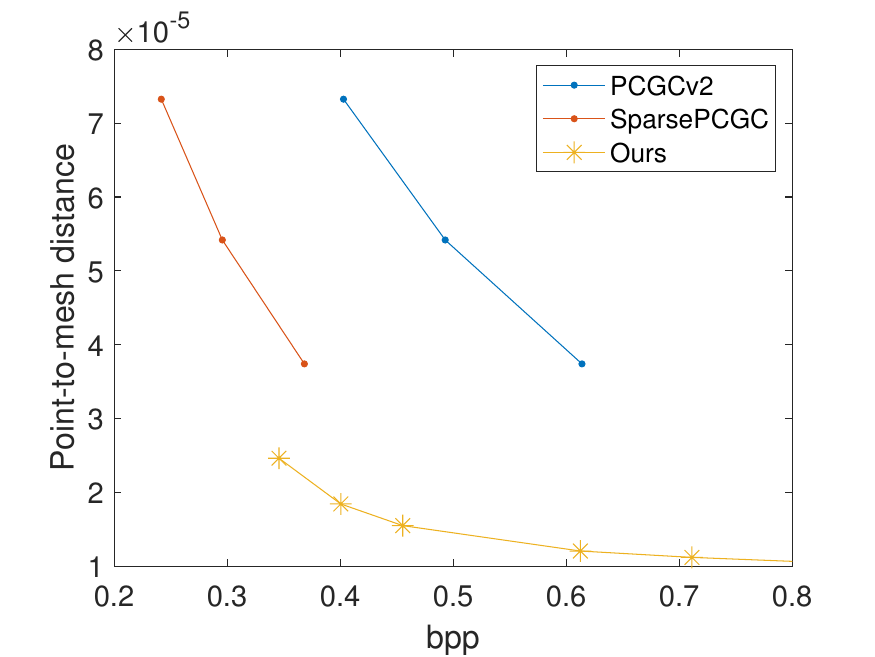}}%
	\hfil
	\subfloat[]{\includegraphics[width=1.6in]{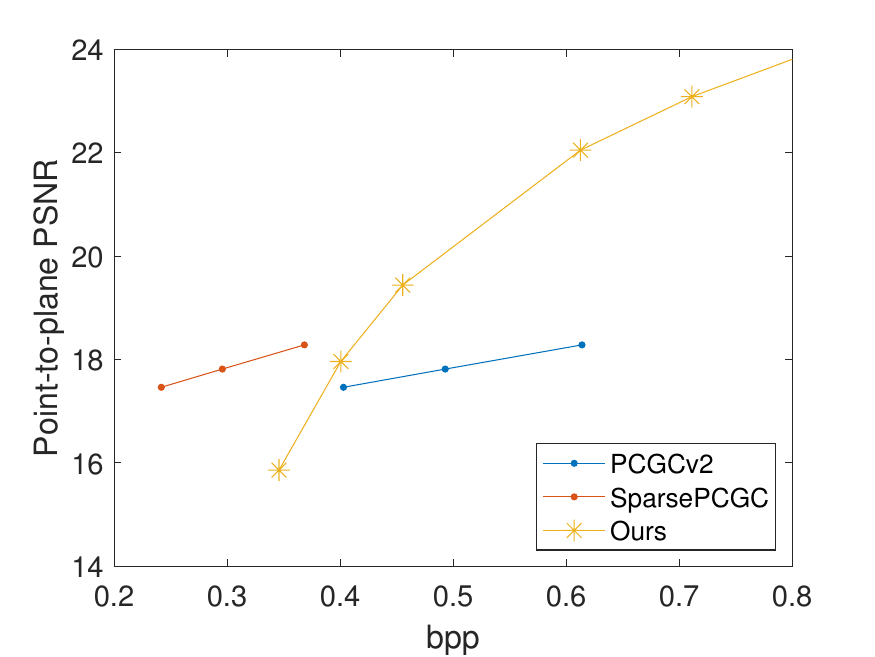}}%
	\caption{Quantitative comparisons with sparse-tensor-based methods.}
	\label{fig:rd_curves_shapenet}
	\end{figure}

	In \cref{fig:rd_curves_shapenet}, we compare our method with PCGCv2 and SparsePCGC and showcase the R-D curves evaluated on the ShapeNet~\cite{chang2015shapenet} dataset. We utilize the official checkpoint of PCGCv2 to evaluate its performance. As the implementation of SparsePCGC is unavailable, we provide the results of PCGCv2 with a 40\% bitrate gain under the same distortion as an estimate of the performance of SparsePCGC.
	While our method trails SparsePCGC in PSNR at low bitrates (below 0.4 bpp), the slope of our curves suggests the potential for achieving higher PSNR scores at higher bitrates. Additionally, our method achieves superb rate-distortion performance in terms of the other two metrics. 
	\section{Comparisons on additional datasets}
\begin{figure}
		\centering
		\subfloat[]{\includegraphics[width=1.6in]{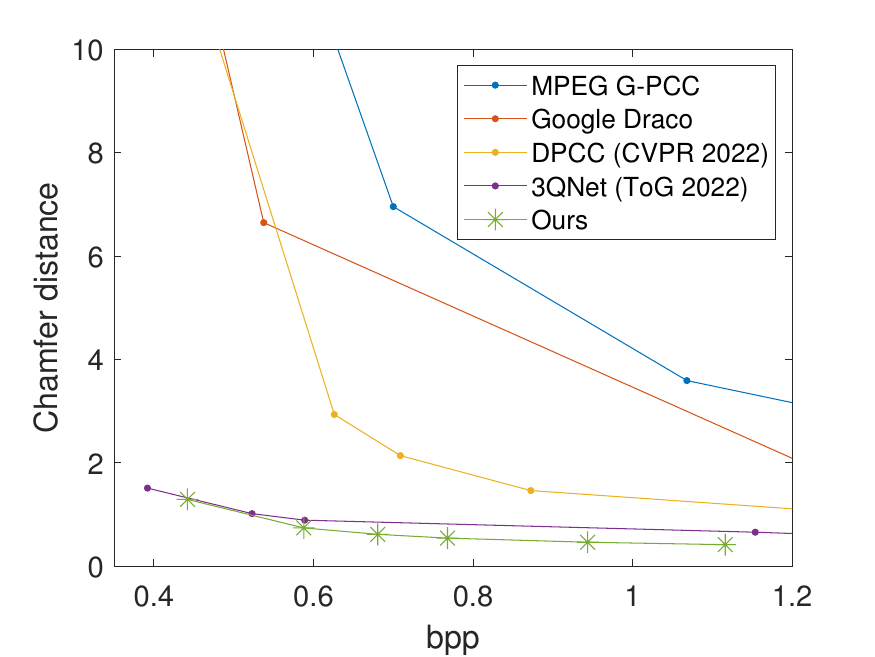}}%
		\hfil
		\subfloat[]{\includegraphics[width=1.6in]{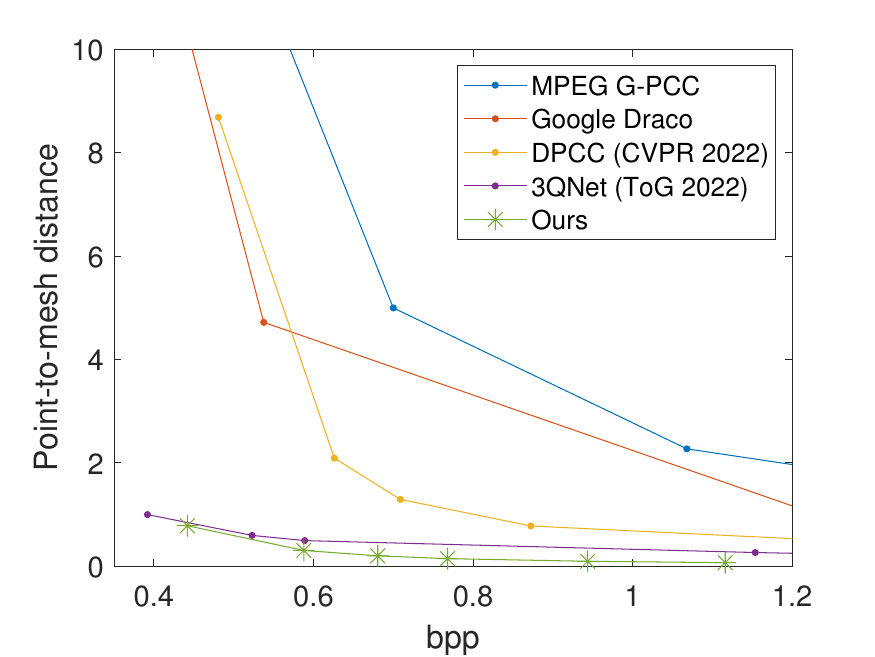}}%
		\hfil
		\subfloat[]{\includegraphics[width=1.6in]{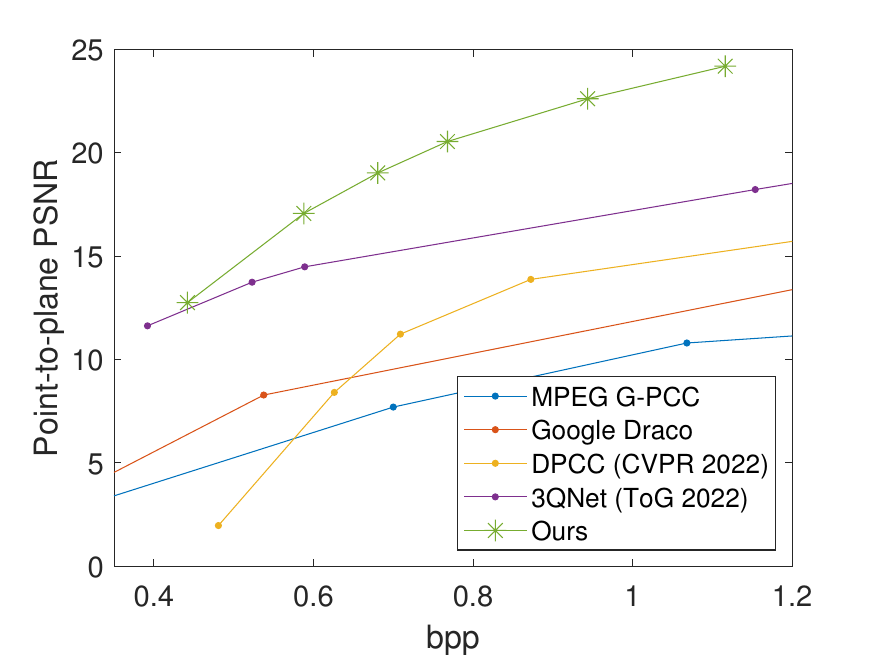}}%
		\caption{Quantitative results on the Florence Superface dataset.}
		\label{fig:rd_curves_face}
	\end{figure}

\begin{figure}
		\centering
		\subfloat[]{\includegraphics[width=1.6in]{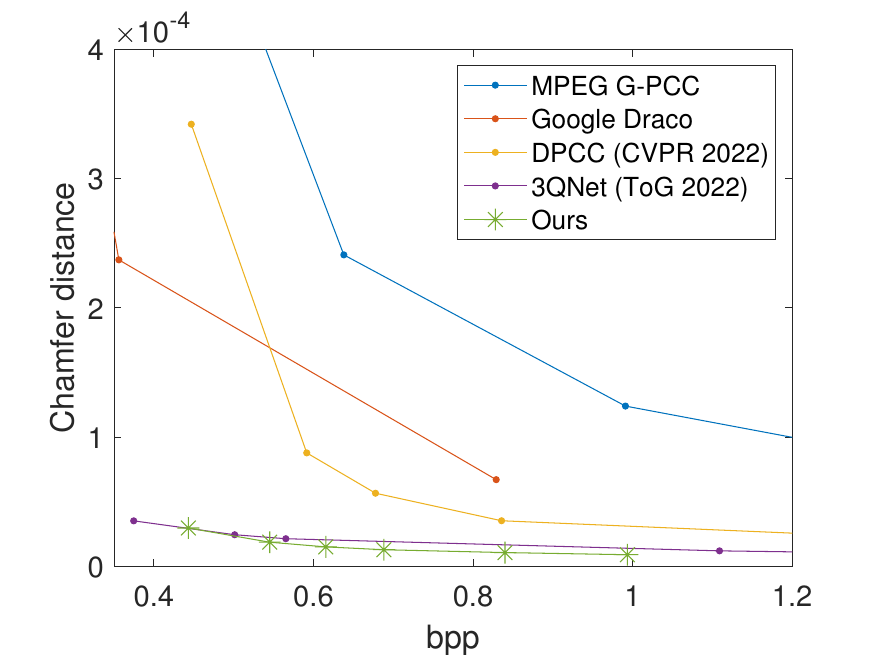}}%
		\hfil
		\subfloat[]{\includegraphics[width=1.6in]{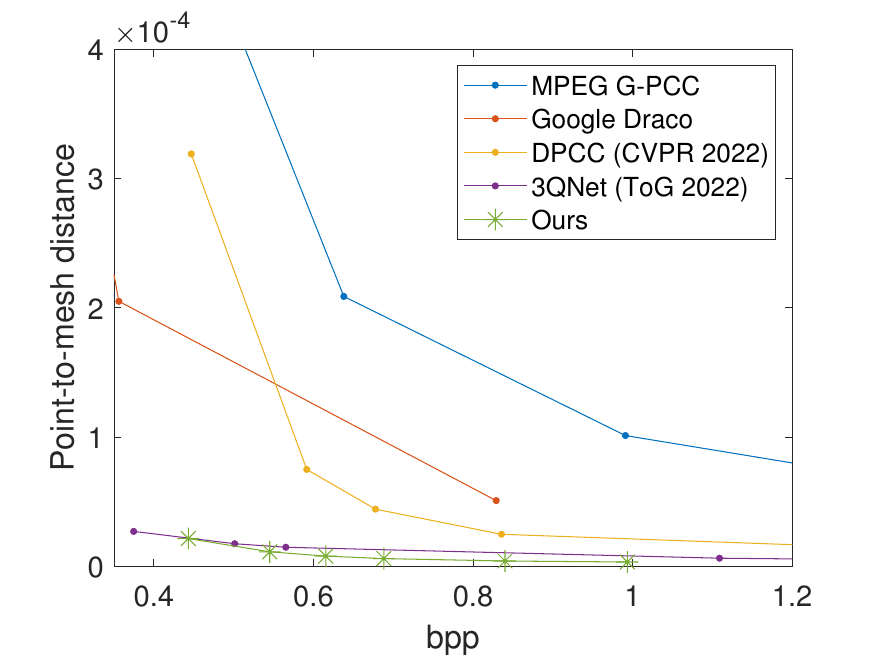}}%
		\hfil
		\subfloat[]{\includegraphics[width=1.6in]{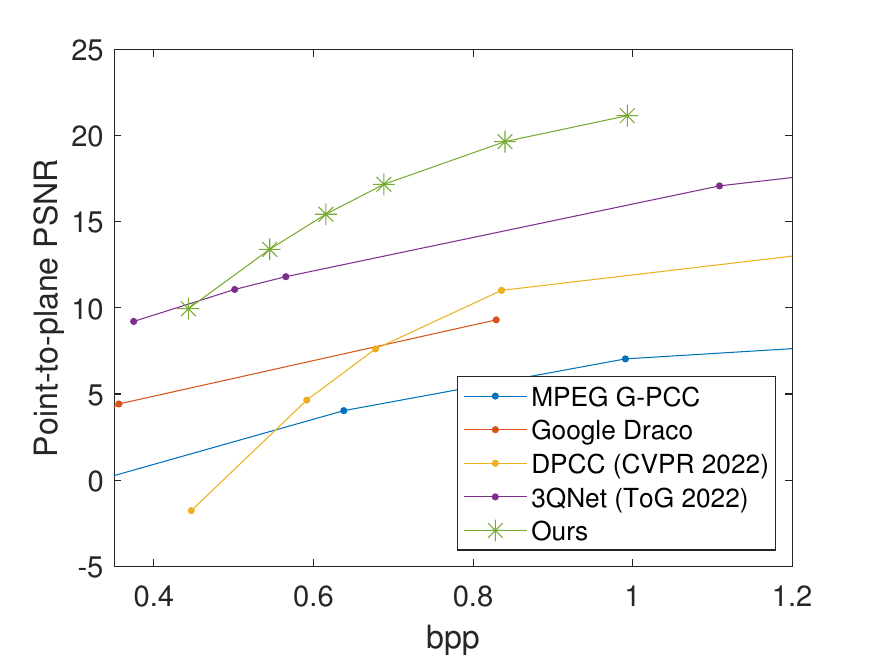}}%
		\caption{Quantitative results on the MPI Dynamic FAUST dataset.}
	\label{fig:rd_curves_dfaust}
	\end{figure}

\begin{figure}
		\centering
		\includegraphics[width=0.65\linewidth]{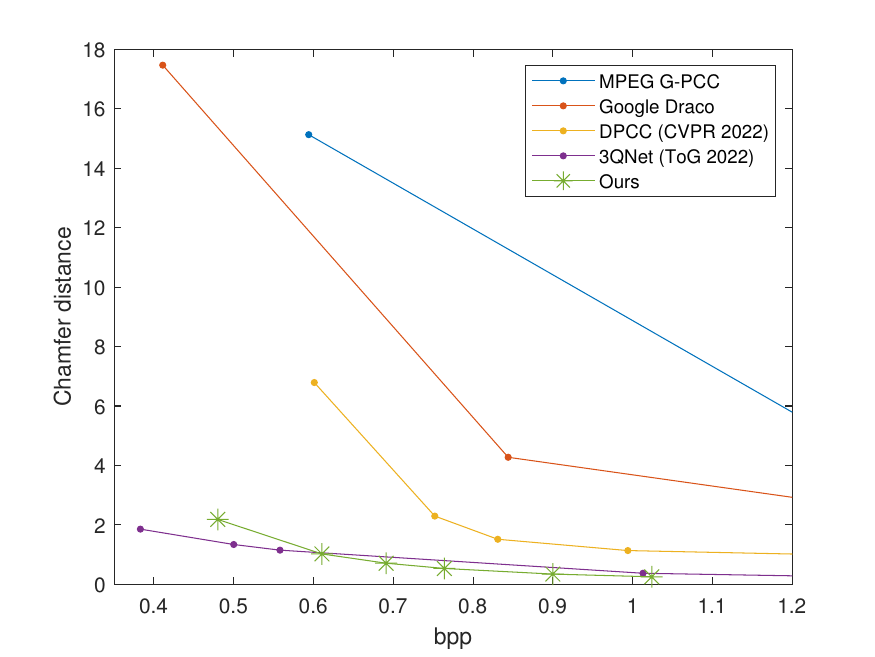}
		\caption{Quantitative results on the Ford dataset.}
		\label{fig:rd_curves_ford}
	\end{figure}

	We evaluate the rate-distortion performance on the Florence Superface~\cite{DBLP:conf/eccv/BerrettiBP12} dataset and the MPI Dynamic FAUST~\cite{dfaust:CVPR:2017} dataset. The former contains 20 3D faces, while the latter consists of 200 3D human bodies, both of which are semantically different from our training data. For each shape in these two datasets, we densely sample its mesh representation at $120$k points to generate the point cloud. We provide R-D curves tested on them in \cref{fig:rd_curves_face} and \cref{fig:rd_curves_dfaust} respectively.
	Additionally, in \cref{fig:rd_curves_ford}, we present the performance evaluated on real-scanned LiDAR point clouds from a subset of the scene-level Ford dataset~\cite{pandey2011ford}.
	Despite our training data only contains man-made CAD models, our method still maintains superiority in compressing those unseen point clouds.
	
\begin{figure}
		\centering
		\includegraphics[width=\linewidth]{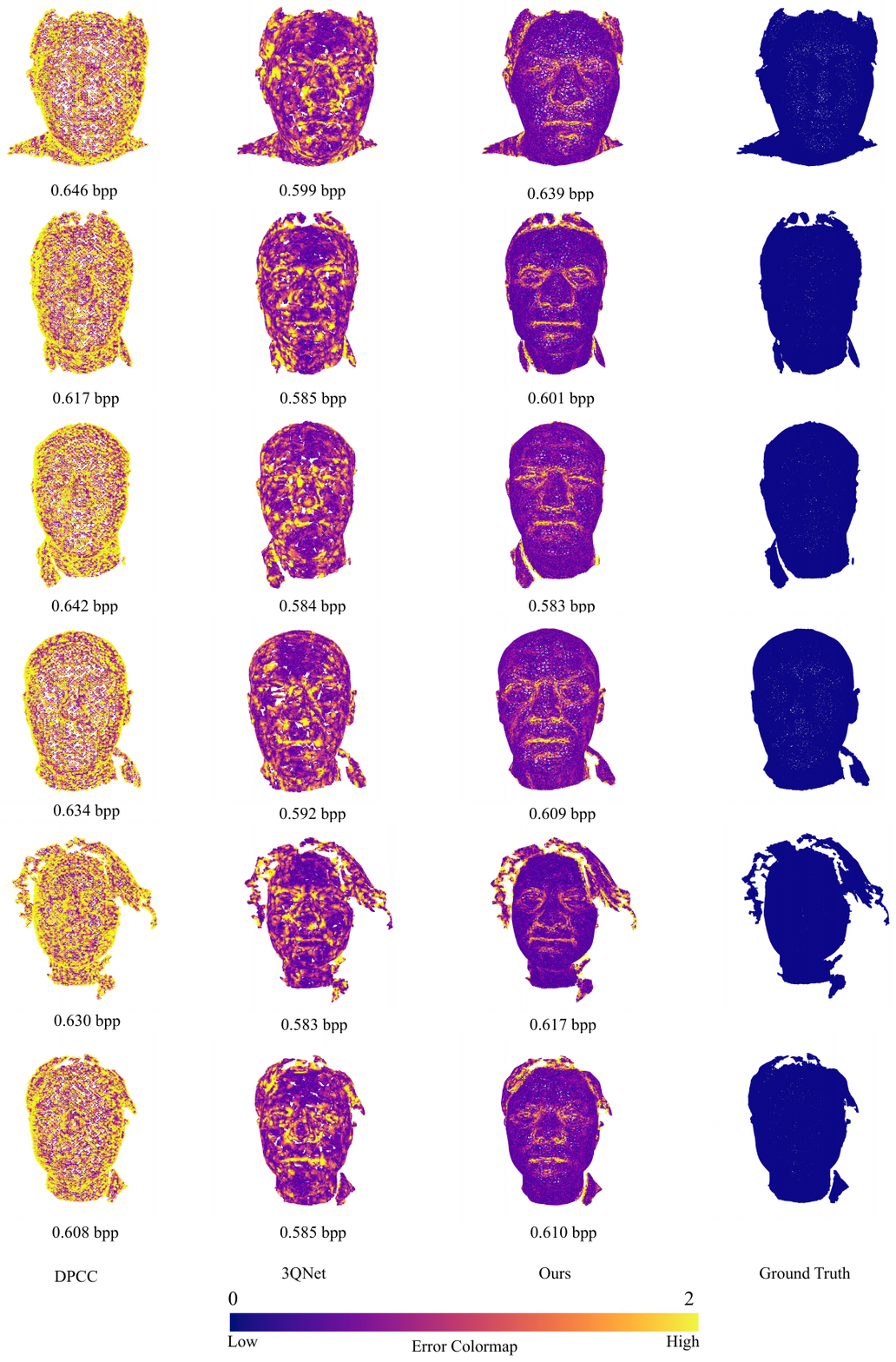}
		\caption{Perceptual results on the Florence Superface dataset.}
		\label{fig:faces}
	\end{figure}

\begin{figure}
		\centering
		\includegraphics[width=\linewidth]{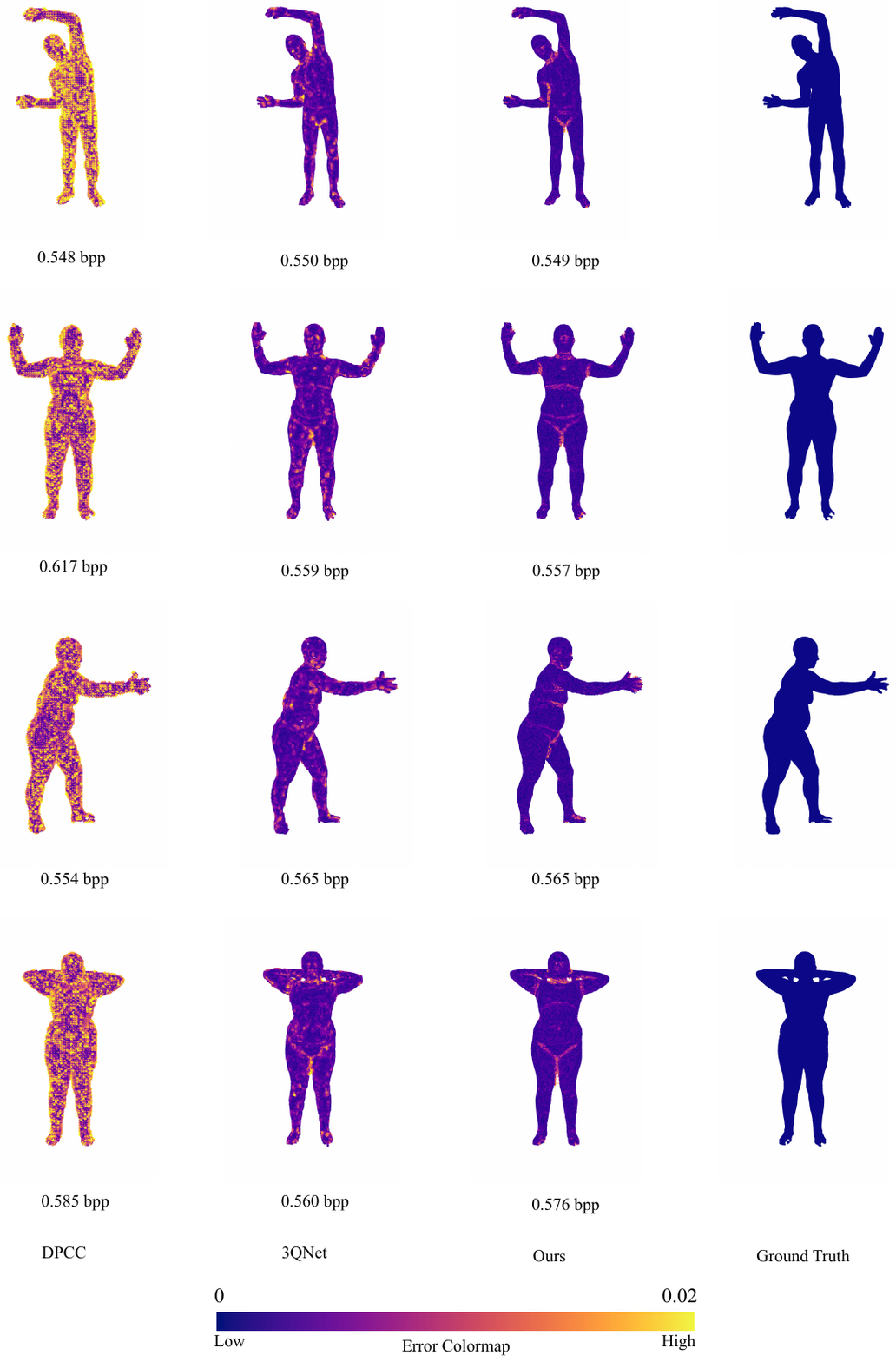}
		\caption{Perceptual results on the MPI Dynamic FAUST dataset.}
		\label{fig:body}
	\end{figure}

\begin{figure}
		\centering
		\includegraphics[width=\linewidth]{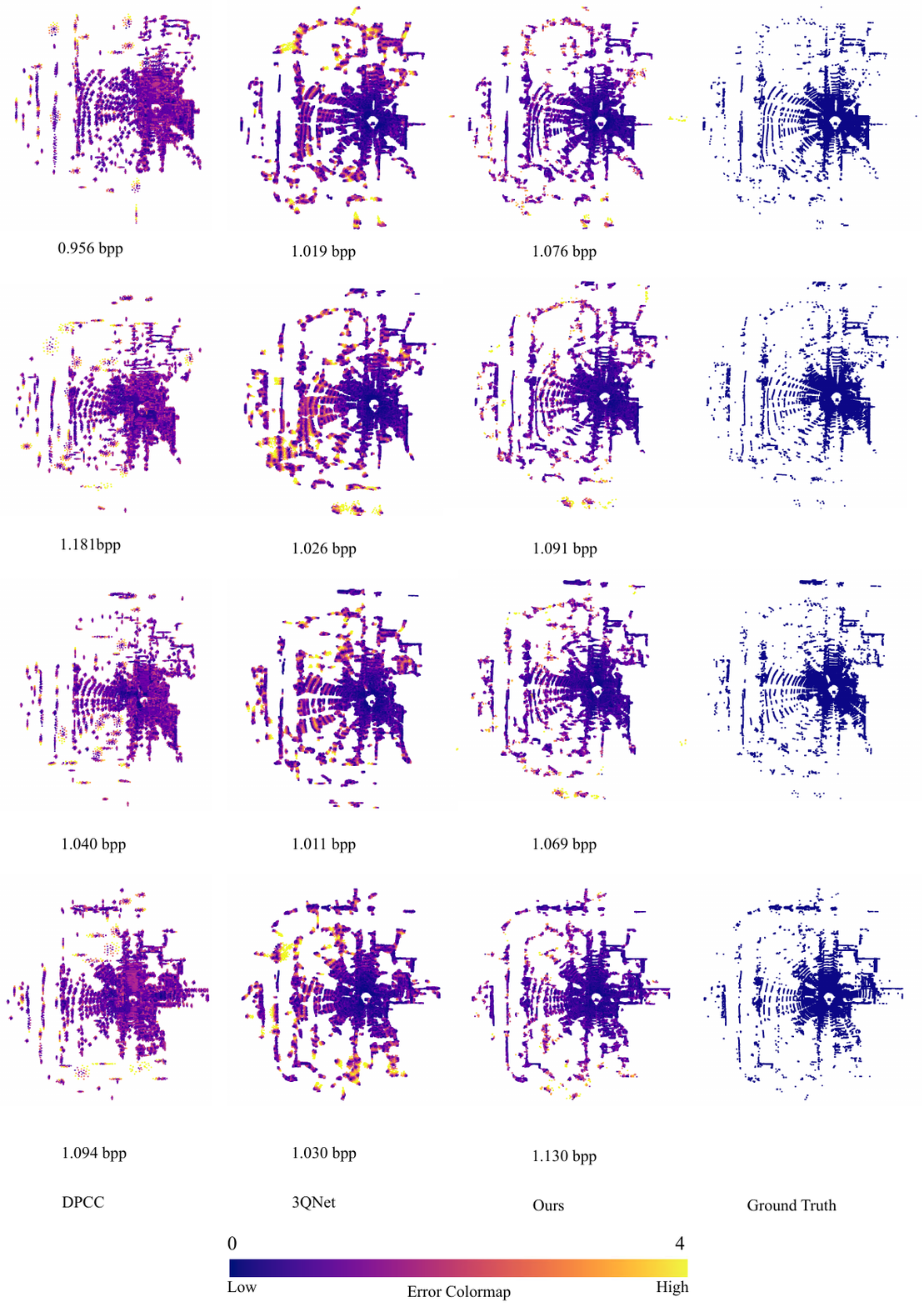}
		\caption{Perceptual results on the Ford dataset.}
		\label{fig:lidar}
	\end{figure}

	We present additional perceptual results on point cloud geometry compression in \cref{fig:faces}, \cref{fig:body} and \cref{fig:lidar}. The error colormap demonstrates clearly superior performance of the proposed method to other competing methods. The more complex the 3D geometry, the more advantageous our method.

	\newpage
	\section{Additional ablation study}
	In \cref{tab:superresolution_abl}, we showcase the efficacy of the proposed compression system in working as an arbitrary-scale upsampling network. This experiment is conducted using the PU-GAN dataset~\cite{li2019pu}, employing the same experimental setup as outlined in the main text. For comparison, we employ Grad-PU~\cite{he2023grad} to upsample the sparse point cloud compressed by our method, serving as a competing baseline. 
	Despite our compression system not being trained explicitly for upsampling, its upsampling performance closely rivals that of a dedicated upsampling network, while also achieving a significant reduction in upsampling runtime. This highlights the effectiveness of our compression method in serving as an arbitrary-scale upsampling network.
	\begin{table}[tb]
		\caption{Comparison in the upsampling performance under various upsampling rates ranging from $\times2$ to $\times8$.\label{tab:superresolution_abl}}
		\centering
		\begin{tabular}{lccclcccl}
			\toprule
			Methods&\multicolumn{3}{c}{Ours (Comp)+Grad-PU~\cite{he2023grad}}&&\multicolumn{3}{c}{Ours (Comp)+Ours (Up)}&\\ \cline{2-4}\cline{6-8}
			\multirow{2}{*}{Up rates}&CD~$\downarrow$    & P2M~$\downarrow$    & Time~$\downarrow$ &           & CD~$\downarrow$    & P2M~$\downarrow$             & Time~$\downarrow$&    \\ 
			&$10^{-5}$&$10^{-5}$&s&&$10^{-5}$&$10^{-5}$&s& \\ \midrule
			$\times2$&\textbf{8.894}&\textbf{4.592}&2.053&&10.922&5.238&\textbf{0.172}&\\
			$\times3$&\textbf{6.594}&\textbf{4.294}&4.226&&7.699&4.756&\textbf{0.173}&\\
			$\times4$&\textbf{5.448}&\textbf{4.156}&7.336&&6.262&4.614&\textbf{0.171}&\\
			$\times5$&\textbf{4.861}&\textbf{4.096}&11.877&&5.678&4.632&\textbf{0.170}&\\
			$\times6$&\textbf{4.511}&\textbf{4.069}&17.118&&5.396&4.695&\textbf{0.173}&\\
			$\times7$&\textbf{4.302}&\textbf{4.076}&23.935&&5.243&4.780&\textbf{0.173}&\\
			$\times8$&\textbf{4.192}&\textbf{4.109}&30.894&&5.171&4.873&\textbf{0.170}&\\
			\bottomrule
		\end{tabular}
	\end{table}
	\section{Visualizing the reconstruction error of upsampled point clouds}
	
\begin{figure}
		\centering
		\includegraphics[width=\linewidth]{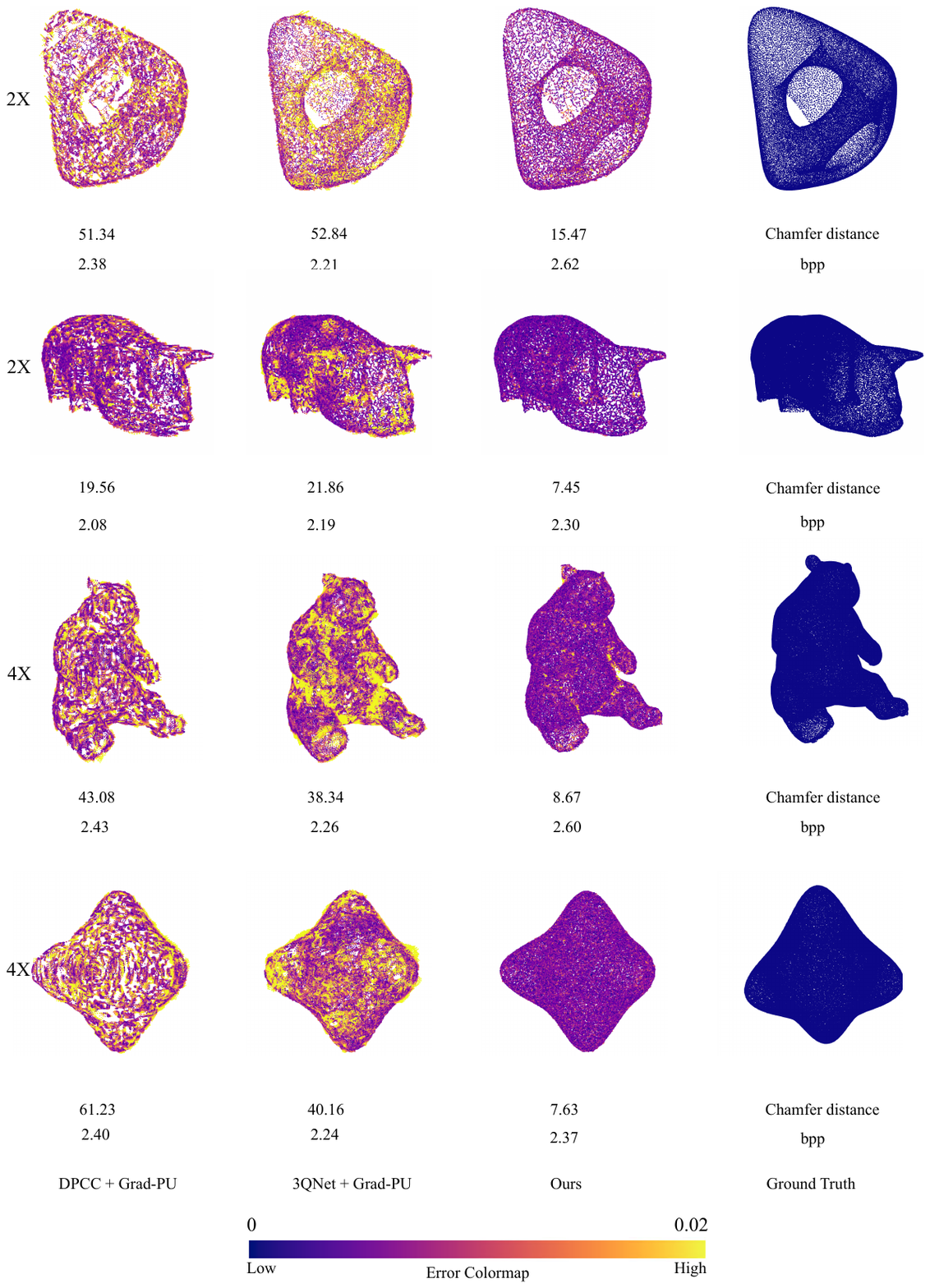}
		\caption{Perceptual results on the PUGAN dataset.}
		\label{fig:pu1}
	\end{figure}

\begin{figure}
		\centering
		\includegraphics[width=\linewidth]{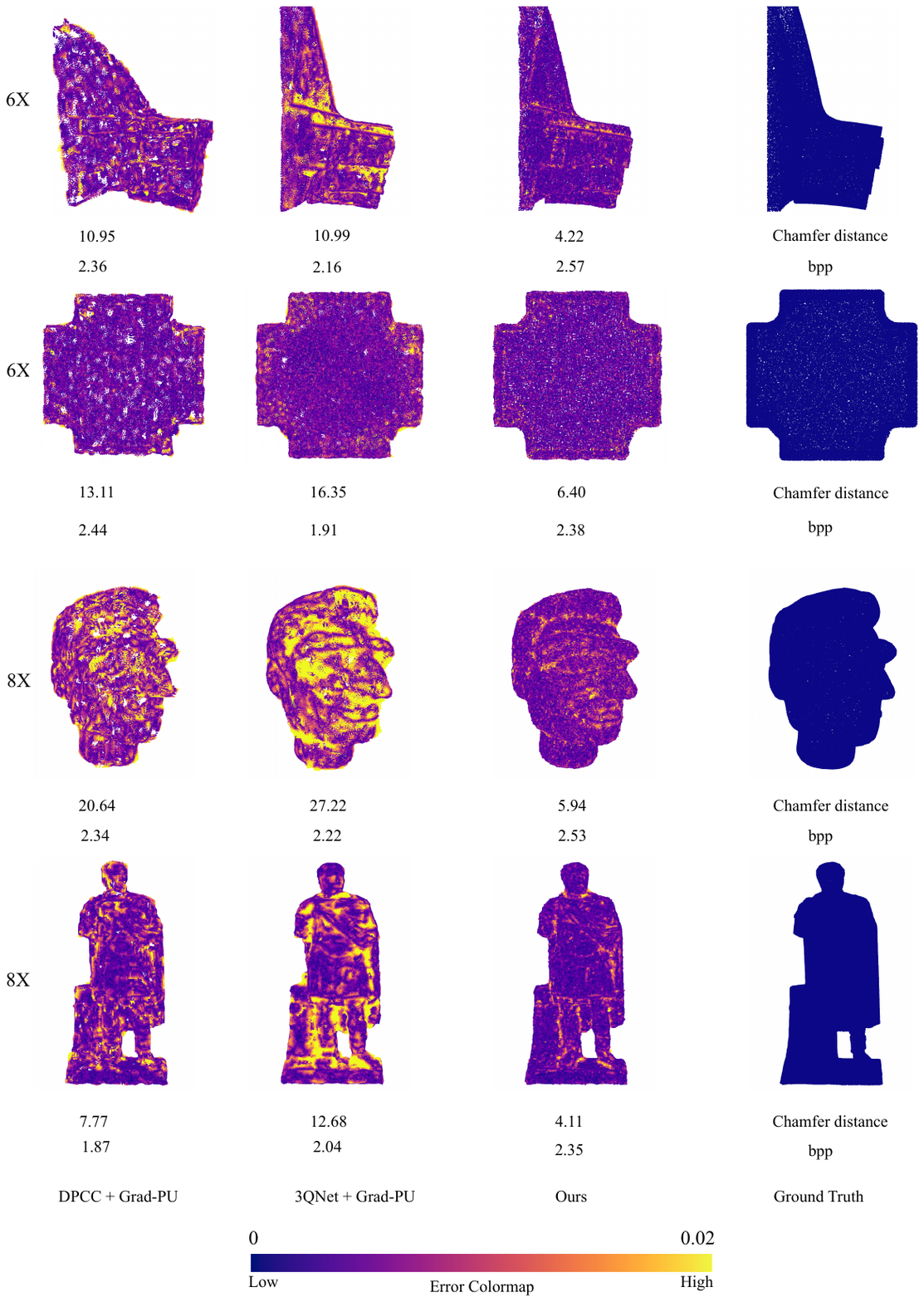}
		\caption{Perceptual results on the PUGAN dataset.}
		\label{fig:pu2}
	\end{figure}

	In Section 4.3 of the main text, we discussed the objective performance of upsampling sparse decompressed point clouds. These sparse point clouds were compressed at near bitrates using various compression methods and subsequently upsampled at the upsampling rates ranging from $\times2$ to $\times8$. Here, we present some perceptual results of upsampling sparse decompressed point clouds in \cref{fig:pu1} and \cref{fig:pu2}. For each presented object, we also provide the bitrate to compress the sparse point cloud, along with the Chamfer distance (CD) between the upsampled point cloud and ground truth. The unit of CD is $10^{-5}$.

\end{document}